\documentclass[letterpaper]{article} % DO NOT CHANGE THIS
\usepackage{aaai23}  % DO NOT CHANGE THIS
\usepackage{times}  % DO NOT CHANGE THIS
\usepackage{helvet}  % DO NOT CHANGE THIS
\usepackage{courier}  % DO NOT CHANGE THIS
\usepackage[hyphens]{url}  % DO NOT CHANGE THIS
\usepackage{graphicx} % DO NOT CHANGE THIS
\usepackage{natbib}  % DO NOT CHANGE THIS AND DO NOT ADD ANY OPTIONS TO IT
\usepackage{caption} % DO NOT CHANGE THIS AND DO NOT ADD ANY OPTIONS TO IT
\usepackage{cite}
\usepackage{color}
\usepackage{multirow}
\usepackage{subfigure}
\usepackage[linesnumbered, ruled]{algorithm2e}
\usepackage{xspace}
\usepackage{dsfont}
\usepackage{epsfig}
\usepackage{epstopdf}
\usepackage{booktabs}
\usepackage{amsmath}
\usepackage{amsthm}
\usepackage{amsfonts}
\usepackage{bm}

\newtheorem{mydef}{Definition}

\newcommand{\bitem}[1]{\noindent$\bullet$ \textbf{#1}}

\newcommand{\ie}{\emph{i.e.,}\xspace}

\newcommand{\eg}{\emph{e.g.,}\xspace}

\newcommand{\name}{ST-SSL\xspace}
\newcommand{\figureautorefname}{Fig.}
\newcommand{\tableautorefname}{Tab.}

\DeclareMathOperator{\wh}{where}

\title{Spatio-Temporal Self-Supervised Learning for Traffic Flow Prediction }

\author {
    Jiahao Ji\textsuperscript{\rm 1},
    Jingyuan Wang\textsuperscript{\rm 1,\rm 2,\rm 3}\thanks{Corresponding author: jywang@buaa.edu.cn},
    Chao Huang\textsuperscript{\rm 4},\\
    Junjie Wu\textsuperscript{\rm 3},
    Boren Xu\textsuperscript{\rm 1},
    Zhenhe Wu\textsuperscript{\rm 1},
    Junbo Zhang\textsuperscript{\rm 5,\rm 6},
    Yu Zheng\textsuperscript{\rm 5,\rm 6}
}
\affiliations {
    \textsuperscript{\rm 1}School of Computer Science \& Engineering, Beihang University, China\\
    \textsuperscript{\rm 2}Peng Cheng Laboratory, China\quad 
    \textsuperscript{\rm 3}School of Economics \& Management, Beihang University, China\\
    \textsuperscript{\rm 4}Department of Computer Science, Musketeers Foundation Institute of Data Science, University of Hong Kong, China\\
    \textsuperscript{\rm 5}JD Intelligent Cities Research, Beijing, China\quad
    \textsuperscript{\rm 6}JD iCity, JD Technology, Beijing, China%\\
}

\begin{document}

\maketitle

\begin{abstract}
    Robust prediction of citywide traffic flows at different time periods plays a crucial role in intelligent transportation systems. While previous work has made great efforts to model spatio-temporal correlations, existing methods still suffer from two key limitations: $i$) Most models collectively predict all regions' flows without accounting for spatial heterogeneity, \ie different regions may have skewed traffic flow distributions. $ii$) These models fail to capture the temporal heterogeneity induced by time-varying traffic patterns, as they typically model temporal correlations with a shared parameterized space for all time periods. To tackle these challenges, we propose a novel Spatio-Temporal Self-Supervised Learning (\name\footnote{The paper was done when Jiahao Ji was an intern at JD Intelligent Cities Research under the supervision of Junbo Zhang (msjunbozhang@outlook.com).}) traffic prediction framework which enhances the traffic pattern representations to be reflective of both spatial and temporal heterogeneity, with auxiliary self-supervised learning paradigms. Specifically, our \name is built over an integrated module with temporal and spatial convolutions for encoding the information across space and time. To achieve the adaptive spatio-temporal self-supervised learning, our \name first performs the adaptive augmentation over the traffic flow graph data at both attribute- and structure-levels. On top of the augmented traffic graph, two SSL auxiliary tasks are constructed to supplement the main traffic prediction task with spatial and temporal heterogeneity-aware augmentation. Experiments on four benchmark datasets demonstrate that \name consistently outperforms various state-of-the-art baselines. Since spatio-temporal heterogeneity widely exists in practical datasets, the proposed framework may also cast light on other spatial-temporal applications. Model implementation is available at \url{https://github.com/Echo-Ji/ST-SSL}.
    % . It is a self-supervised learning-based traffic flow prediction framework that consists of three components: $i$) A spatio-temporal encoder to generate region embeddings from the input traffic data. $ii$) An adaptive data augmentation approach to constructing traffic augmentation that removes data heterogeneity in a data-driven manner. $iii$) Two self-supervised tasks take the augmented and original data as input. Then, they learn the spatial heterogeneity by a clsuter assignment prediction task and model the temporal heterogeneity by a traffic trend classification task, respectively. 
    
    % learns spatial and temporal heterogeneity through two self-supervised tasks: $i$) The first task aims to model spatial heterogeneity by clustering regions with highly dependent flow distributions together and transferring more information among them. $ii$) The second task is to model temporal heterogeneity by encouraging the model to carry temporal information present in all regions, thus enabling customized modeling for different time periods. Furthermore, a data-driven augmentation approach is proposed to prepare data for these two tasks. It extracts global semantic information to adaptively augment different traffic flow data. 
    % Experiments on four traffic benchmark datasets demonstrate that HeST consistently outperforms state-of-the-art baselines for different types of regions and different time periods.
\end{abstract}

\section{Introduction}\label{sec:intro}

% Reliable traffic flow prediction across different spatial regions at different time periods is crucial for the development of smart cities~\cite{pan2019matrix, zhang2021traffic}. For example, accurately predicting traffic flow in popular regions during rush hours can help governments implement traffic controls or send out warnings to maintain public safety. Also, accurate traffic flow prediction for less popular regions at regular times or nighttimes can help transportation departments optimize scheduling strategies for public transportation and reduce operating costs. 
Robust traffic flow prediction across different spatial regions at different time periods is crucial for advancing intelligent transportation systems~\cite{zhang2020spatial}. For example, accurate traffic prediction results can not only enable effective traffic controls in a timely manner, but also mitigate tragedies caused by the sudden traffic flow spike. In general, traffic prediction aims to forecast the traffic volume (\eg inflow and outflow of each region at a given time), from past traffic observations. Recent advances have significantly boosted the research of traffic flow prediction with various deep learning techniques, \eg convolutional neural networks over region grids~\cite{zhang2017deep}, graph neural networks for spatial dependency modeling~\cite{zhang2021traffic}, and attention mechanism for spatial information aggregation~\cite{zheng2020gman}. Although significant efforts have been made to improve the traffic flow prediction results, existing models still face two key shortcomings.

% yu2018spatio

% For example, accurately predicting traffic flow can help governments implement traffic controls or send out warnings to maintain public safety. Generally, traffic prediction aims to forecast the traffic volume (\eg, inflow and outflow of each region at given time), from past traffic observations. Recent advances in deep learning have boosted the research of traffic flow prediction~\cite{zhang2017deep, yao2019revisiting, ji2022stden}. Generally, these models \emph{collectively} predict flows in all regions for any given time through powerful deep networks, \eg convolutional neural network (CNN) and graph neural network (GNN) for spatial dependency modeling, and recurrent neural network (RNN) and temporal convolutional network (TCN) for temporal correlation learning. Although significant improvements have been made in traffic flow prediction, these models still face two key shortcomings.

\begin{figure}[t]
    \centering
    \includegraphics[width=0.98\columnwidth]{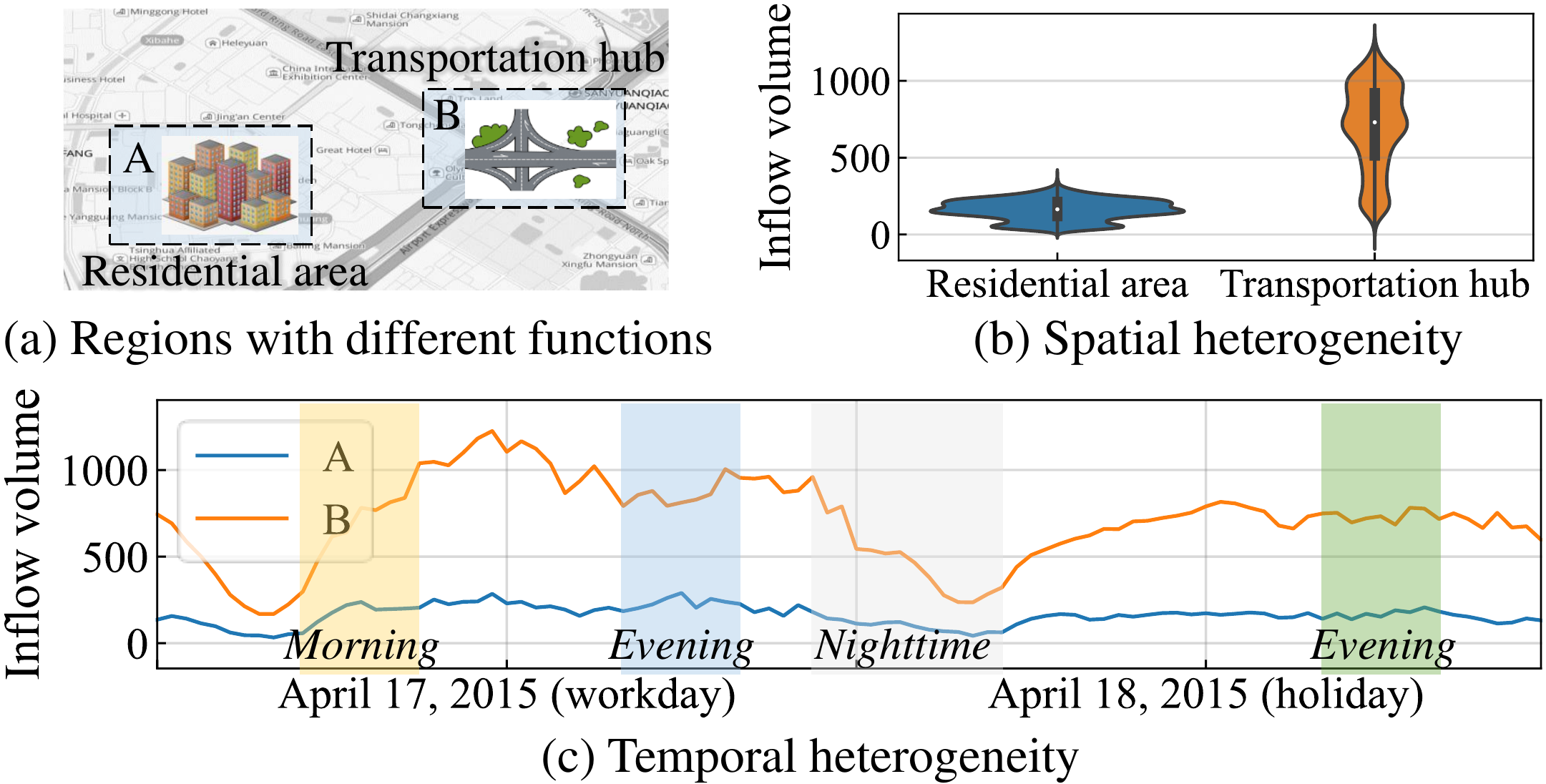}\vspace{-0.2cm}
    \caption{Illustration of our motivation, \ie the spatial and temporal heterogeneity of traffic flow data.}\vspace{-0.5cm}\label{fig:motivation}
\end{figure}

The first limitation is the \emph{lack of modeling spatial heterogeneity} exhibited with skewed traffic distributions across different regions. Taking \figureautorefname{~\ref{fig:motivation}}(a) for example, A and B are two real-world regions in Beijing with different urban functions, namely the residential area and transportation hub. We can observe their quite different traffic flow distributions from \figureautorefname{~\ref{fig:motivation}}(b). However, most existing models ignore such spatial heterogeneity and are easily biased towards popular regions with higher traffic volume, which make them insufficient to learn quality citywide traffic pattern representations. While some studies attempt to capture the heterogeneous flow distributions with multiple parameter sets over different regions~\cite{pan2019matrix, bai2020adaptive}, the involved large parameter size may lead to the suboptimal issue over the skewed-distributed traffic data. Worse still, the high computational and memory cost of these methods make them infeasible to handle large-scale traffic data in practical urban scenarios. In addition, meta-learning has been used in recent approaches~\cite{pan2019urban, ye2022meta} to consider the difference of region traffic distributions. However, the effectiveness of those models largely relies on the collected handcrafted region spatial characteristics, \eg nearby points of interest and density of road networks, which limits the model representation generalization ability.

% that exists in different regions. Taking \figureautorefname{~\ref{fig:motivation}}(a) for example, A, B are two real-world regions with different latent functions, corresponding to a residential area and a transportation hub, respectively. Their flow distributions are quite heterogeneous (see \figureautorefname{~\ref{fig:motivation}}(b)). However, most existing models collectively predict all regions' flows without consideration of spatial heterogeneity, leading to limited prediction accuracy, in particular for less popular regions. Some works have already attempted to model spatial heterogeneity. Multiple models~\cite{yuan2018hetero} and multiple sets of parameters~\cite{pan2019matrix, bai2020adaptive} for different regions are introduced to capture heterogeneous flow distributions. However, these methods may face overfitting problem due to the large growth of model parameters. Besides, they suffer from high memory and computational costs. Meta learning~\cite{pan2019urban, ye2022meta} can also model spatial heterogeneity by generating different initial parameters using static spatial characteristics of a region, \eg nearby points of interests and density of road networks. However, detailed spatial characteristics of a region are not always available. 

Furthermore, current traffic prediction methods model the temporal dynamics with a shared parameter space for all time periods, which can hardly precisely preserve the \emph{temporal heterogeneity} in the latent embedding space. In real-life scenarios, traffic patterns of different regions vary over time, \eg from morning to evening, which results in the temporal heterogeneity as shown in \figureautorefname{~\ref{fig:motivation}}(c). Nevertheless, the parameter space differentiation strategy adopted in~\cite{song2020spatial, li2021spatial} assumes that the temporal heterogeneity is static across the entire time periods, which is not always held, \eg evening traffic patterns can be significantly different for workdays and holidays shown in \figureautorefname{~\ref{fig:motivation}}(c).

To effectively model both spatial and temporal heterogeneity, we present a novel \underline{S}patio-\underline{T}emporal \underline{S}elf-\underline{S}upervised \underline{L}earning framework for predicting traffic flow. To encode spatial-temporal traffic patterns, our \name is built over a graph neural network which integrates temporal and spatial convolutions for information aggregation. To capture the spatial heterogeneity, we design a spatial self-supervised learning paradigm to augment the traffic flow graph at both data-level and structure-level, which is adaptive to the heterogeneous region traffic distributions. Then, the auxiliary self-supervision with a soft clustering paradigm is introduced to be aware of the diverse spatial patterns among different regions. To inject the temporal heterogeneity into our latent representation space, we empower \name to maintain dedicated representations of temporal traffic dynamics with temporal self-supervised learning paradigm. We summarize the key contributions of this work as follows:

\begin{itemize}
    \item To the best of our knowledge, we are the first to propose a novel self-supervised learning framework to model spatial and temporal heterogeneity in traffic flow prediction. This paradigm may shed light on other practical spatio-temporal applications, such as air quality prediction.
    
    % \item We develop an adaptive data augmentation method that can augment spatio-temporal data under the guidance of semantic similarity extracted from the input data. It can improve the quality of augmented data compared with random augmentation.
    
    \item We propose an adaptive heterogeneity-aware data augmentation scheme over the graph-structured spatial-temporal graph against the noise perturbation. 
    
    % \item We propose two self-supervised tasks to learn the spatial and temporal heterogeneity without significantly increasing the parameters of the model. It can be easily transferred to other spatio-temporal models as well.
    
    \item Two self-supervised learning tasks are incorporated to supplement the main traffic prediction task by enforcing the model discrimination ability with the awareness of both spatial and temporal traffic heterogeneity. 
    
    % \item Extensive experiments are conducted on four real-world public datasets. The results show that our model consistently outperforms baselines, especially for the less popular regions and time periods with sparse traffic flow data.
    
    \item Extensive experiments are conducted on four real-world public datasets to show the consistent performance superiority achieved by our \name across various settings.
    
\end{itemize}

% \noindent$\bullet$ To the best of our knowledge, we are the first to propose a novel self-supervised learning framework to model spatial and temporal heterogeneity in traffic flow prediction. This modeling paradigm may shed light on other practical spatio-temporal applications, such as air quality prediction.

% \noindent$\bullet$ We propose an adaptive heterogeneity-aware data augmentation scheme over the graph-structured spatial-temporal graph against the noise perturbation. 

% \noindent$\bullet$ Two self-supervised learning tasks are incorporated to supplement the main traffic prediction task by enforcing the model discrimination ability with the awareness of both spatial and temporal traffic heterogeneity.

% \noindent$\bullet$ Extensive experiments are conducted on four real-world public datasets to show the consistent performance superiority achieved by our \name across various settings.

% \begin{figure}[t]
%     \centering
%     \includegraphics[width=0.95\columnwidth]{figures/paradigm.pdf}\vspace{-0.2cm}
%     \caption{Conceptual comparison of two heterogeneity modeling paradigms in (b) and (c). They both extend from the basic paradigm of traffic flow prediction shown in (a).}\vspace{-0.3cm}\label{fig:paradigm}
% \end{figure}

\section{Preliminaries}\label{sec:pre}
% In this section, we begin with some key definitions. Then, we present our studied task of traffic flow prediction.

% In this section, we begin with key definitions and notations. Then, we formally state our studied traffic prediction task.

\begin{mydef}[Spatial Region]
We partition a city into $N = I \times J$ disjoint geographical grids, in which each grid is considered as a spatial region $r_n (1 \le n \le N)$. We use $\mathcal{V} = \{r_1, \dots, r_N\}$ to denote the spatial region set in a city. 
% grids, in which each grid is regarded as a spatial region $r_n$ ($n \in [1, \dots, N], \wh N = I * J$). $r_n$ is the traffic prediction unit.
\end{mydef}

% \begin{mydef}[Traffic Flow Tensor]
% We split the whole time period (\eg one month) into
% $T$ equal-length continuous time intervals. The citywide start/end traffic flow distribution across spatial regions can be represented as $X \in \mathbb{R}^{N \times T \times 2}$, where each item $x_{i, t} = [x^{s}_{i, t}, x^{e}_{i, t}]$ denotes the $[start, end]$ traffic volume of region $i$ at time interval $t$. 
% \end{mydef}

% \textbf{Problem Definition}. Given the traffic flow data until time interval $t$, the traffic prediction problem aims to predict the start and end traffic flow volume at time interval $t+1$.

\begin{mydef}[Traffic Flow Graph (TFG)]
A traffic flow graph is defined as $\mathcal{G} = \left(\mathcal{V}, \mathcal{E}, \bm A, \mathcal{X}_{t-T:t}\right)$, where $\mathcal{V}$ is the set of spatial regions (nodes) with the size of $|\mathcal{V}| = N$, and $\mathcal{E}$ is a set of edges connecting two spatially adjacent regions in $\mathcal{V}$. The adjacent matrix of our traffic flow graph is denoted as $\bm A \in \mathbb{R}^{N \times N}$. We represent the citywide traffic inflow and outflow data over previous $T$ time steps with a traffic tensor $\mathcal{X}_{t-T:t} \in \mathbb{R}^{T \times N \times 2} = \left(\bm X_{t-T}, \dots, \bm X_{t} \right)$. The traffic volume information of all regions $\mathcal{V}$ at the $t$-th time slot is denoted as $\bm X_{t} \in \mathbb{R}^{N \times 2}$. 
% and $\bm A \in \mathbb{R}^{N \times N}$ is the binary adjacency matrix describing region relations. $\mathcal{X}_{t-T:t} = \left(\bm X_{t-T}, \dots, \bm X_{t} \right)$ is a traffic flow sequence, where $T$ is the time window size and $\bm X_{t} \in \mathbb{R}^{N \times 2}$ represents inflow and outflow of all regions observed at time step $t \in \mathbb{N}$.
\end{mydef}

% \textbf{Problem Definition}. Given TFG data until time step $t$, the traffic flow prediction problem aims to predict the traffic flow at time step $t+1$, \ie $\bm X_{t+1} \in \mathbb{R}^{N \times 2}$.

\noindent \textbf{Problem Statement}. Given the historical traffic flow graph $\mathcal{G}$ till the current time step, we aim to learn a predictive function which accurately estimates the traffic volume of all regions at the future time step $t+1$, \ie $\bm X_{t+1} \in \mathbb{R}^{N \times 2}$.

\section{Methodology}\label{sec:method}

\begin{figure*}[t]
    \centering
    \includegraphics[width=0.98\textwidth]{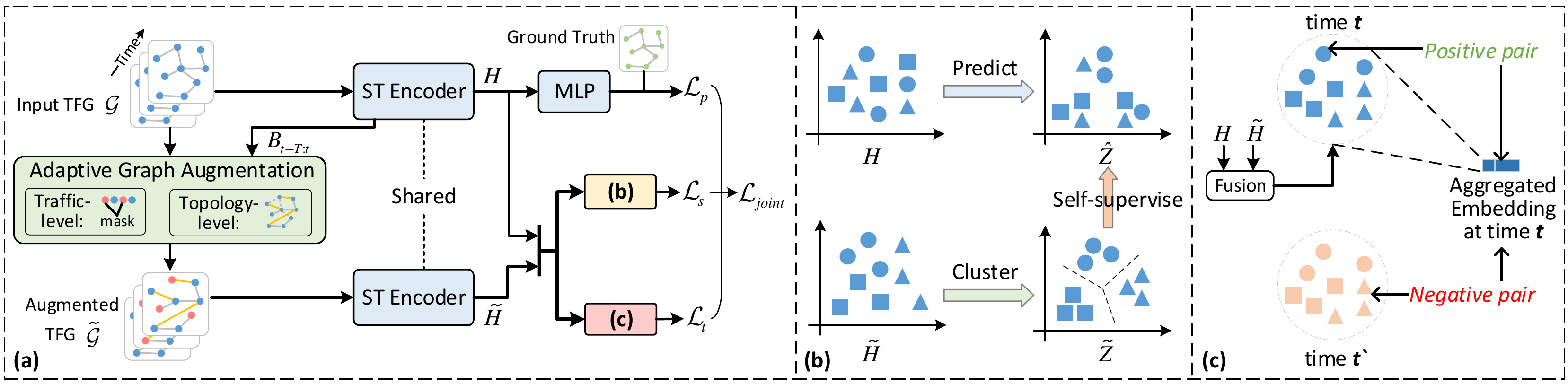}
    \vspace{-0.2cm}
    \caption{(a): The overall architecture of \name. (b): Spatial heterogeneity modeling. (c): Temporal heterogeneity modeling.}
    \vspace{-0.4cm}
    \label{fig:framework}
\end{figure*}

% This section elaborates our proposed \name model shown in \figureautorefname~\ref{fig:framework}. The model consists of three components: spatio-temporal (ST) encoder, adaptive data augmentation (AdaAug), and self-supervised learning (SSL) for heterogeneity modeling. 

This section elaborates on the technical details of our \name model with the overall architecture shown in \figureautorefname~\ref{fig:framework}.

\subsection{Spatio-Temporal Encoder}

We firstly propose a spatio-temporal (ST) encoder to jointly preserve the ST contextual information over the traffic flow graph, so as to jointly model the sequential patterns of traffic data across different time steps and the geographical correlations among spatial regions. Towards this end, we integrate the temporal convolutional component with the graph convolutional propagation network as the backbone for spatial-temporal relational representation.

% Given a TFG $\mathcal{G}$, Spatio-Temporal (ST) encoder aims to learn a region embedding matrix $\bm H$. We introduce a temporal convolution layer (TC) and a spatial convolution layer (SC) to fully integrate temporal and spatial information into the region embedding.

For encoding the temporal traffic patterns, we adopt the 1-D causal convolution along the time dimension with a gated mechanism~\cite{yu2018spatio}. Specifically, our temporal convolution (TC) takes the traffic flow tensor as the input and outputs a time-aware embedding for each region:
\begin{equation}\label{eq:tcl}\small
    \left(\bm{B}_{t-T_{out}}, \dots, \bm{B}_{t}\right) = \mathrm{TC}\left(\bm X_{t-T}, \dots, \bm X_{t}\right),
\end{equation}
\noindent where $\bm{B}_{t} \in \mathbb{R}^{N \times D}$ denotes the region embedding matrix at the time step $t$. The $n$-th row $\bm b_{t,n} \in \mathbb{R}^D$ corresponds to the embedding of region $r_n$. Here, $D$ denotes the embedding dimensionality. $T_{out}$ is the length of the output embedding sequence after convolutional operations in TC encoder.

% Given a traffic flow sequence $\mathcal{X}_{t-T:t} = \left(\bm X_{t-T}, \dots, \bm X_{t} \right)$, TC outputs an embedding sequence 
% \begin{equation}\label{eq:tcl}\small
%     \left(\bm{B}_{t-T_{out}}, \dots, \bm{B}_{t}\right) = \mathrm{TC}\left(\bm X_{t-T}, \dots, \bm X_{t}\right),
% \end{equation}
% where $\bm{B}_{t} \in \mathbb{R}^{N \times D}$ denotes the embedding matrix at time $t$, and the $n$-th row $\bm b_{t,n} \in \mathbb{R}^D$ corresponds to the embedding of region $r_n$. $T_{out}$ is the length of the output sequence. 

For capturing the region-wise spatial correlations, we design our spatial convolution (SC) encoder based on a graph-based message passing mechanism presented as follows:
\begin{equation}\label{eq:scl}\small
    \bm{E}_t = \mathrm{SC}\left(\bm{B}_t, \bm A\right).
\end{equation}
\noindent $\bm A$ is the region adjacency matrix of $\mathcal{G}$. After our SC encoder, we can obtain the refined embeddings $\left(\bm{E}_{t-T_{out}}, \dots, \bm{E}_{t}\right)$ of all regions by injecting the geographical context.

% For each $\bm{B}_{t}$, we employ a graph convolution network~\cite{kipf2017semi} as our SC, whose update is 
% \begin{equation}\label{eq:scl}\small
%     \bm{E}_t = \mathrm{SC}\left(\bm{B}_t, \bm A\right), 
% \end{equation}
% where $\bm A$ is the adjacency matrix of $\mathcal{G}$. Through SC, we obtain a sequence $\left(\bm{E}_{t-T_{out}}, \dots, \bm{E}_{t}\right)$. 

Our ST encoder is built with a ``sandwich'' block structure, in which TC $\to$ SC $\to$ TC is each individual block. By stacking multiple blocks, we can obtain a sequence of embedding matrix $\left(\bm{H}_{t-T'}, \dots, \bm{H}_{t}\right)$ with the temporal dimension of $T'$ after several convolutions. After ST encoder-based embedding propagation and aggregation, the temporal dimension $T'$ reduces to zero and we generate the final embedding matrix $\bm H \in \mathbb{R}^{N \times D}$ for our ST encoder, in which each row $\bm h_n \in \mathbb{R}^D$ denotes the final embedding of region $r_n$. 

In the next subsection, we will perform the adaptive augmentation over the $\left(\bm{B}_{t-T}, \dots, \bm{B}_{t}\right)$ output from the first TC encoder layer (Sec 3.2), and self-supervised learning with the spatial-temporal heterogeneity modeling based on the final region embedding matrix $\bm H$ (Sec 3.3-Sec 3.4).

% We adopt an ``sandwich'' structure, \ie TC $\to$ SC $\to$ TC, as the main block of our ST encoder. By stacking multiple blocks, we obtain a sequence of $\left(\bm{H}_{t-T_{final}}, \dots, \bm{H}_{t}\right)$. In this sequence, we ensure $T_{final}$ is equal to 0, which means the sequence contains only one embedding matrix, \ie $\bm{H}_{t}$. We denote it as $\bm H \in \mathbb{R}^{N \times D}$ for simplicity, and the $n$-th row $\bm h_n \in \mathbb{R}^D$ corresponds to the final embedding of region $r_n$.

% Overall, out ST encoder gives two outputs: output of the first TC layer, \ie $\left(\bm{B}_{t-T_{out}}, \dots, \bm{B}_{t}\right)$ for Section 3.2, and the final embedding matrix $\bm H$ for Sections 3.3 and 3.4.

\subsection{Adaptive Graph Augmentation on TFG}

% The Adaptive data Augmentation (AdaAug) aims to generate TFG augmentation in a data-driven manner. We first measure heterogeneity among different regions. Then, we adaptively perturb the input TFG $\mathcal{G}$ and generate its augmentation $\tilde{\mathcal{G}}$ guided by the heterogeneity measurement.

We devise two phases of graph augmentation schemes on TFG $\mathcal{G} = \left(\mathcal{V}, \mathcal{E}, \bm A, \mathcal{X}_{t-T:t}\right)$ with traffic-level data augmentation and graph topology-level structure augmentation, which is adaptive to the learned heterogeneity-aware region dependencies in terms of their traffic regularities.

\subsubsection{Region-wise Heterogeneity Measurement.} 
For a region $r_n$, its embedding sequence $(\bm{b}_{t-T, n}, \dots, \bm{b}_{t, n})$ within $T$ time steps from rows of $\left(\bm{B}_{t-T}, \dots, \bm{B}_{t}\right)$ is used to generate an overall embedding as:
\begin{equation}\small\label{eq:ta}
    \bm u_n = \sum_{\tau = t-T}^{t} p_{\tau,n}\cdot \bm b_{\tau,n},~\wh~ p_{\tau,n} = \bm b_{\tau,n}^\top \cdot \bm w_0.
\end{equation}
\noindent $\bm u_n$ is the aggregated representation over region $r_n$'s embedding sequence across different time steps based on the derived aggregation weight $p_{\tau,n}$. Here, $\tau$ is the index of the time step range $(t-T, t)$. The aggregation weight $p_{\tau,n}$ reflects the relevance between the time step-specific traffic pattern ($\bm b_{\tau,n}$) and the overall traffic transitional regularities ($\bm u_n$). $\bm b_{\tau,n}$ is region $r_n$'s embedding at time step $\tau$ and $\bm w_0 \in \mathbb{R}^{D}$ is a learnable parameter vector for transformation.

% \subsubsection{Measuring Heterogeneity among Regions.}
% For a region $r_n$, we take its embedding sequence $(\bm{b}_{t-T, n}, \dots, \bm{b}_{t, n})$ as input, and calculate a summary embedding by
% \begin{equation}\small\label{eq:ta}
%     \bm u_n = \sum_{\tau = t-T}^{t} p_{\tau,n}\bm b_{\tau,n},~\wh~ p_{\tau,n} = \bm b_{\tau,n}^\top \bm w_0.
% \end{equation}
% Here, $\bm u_n$ is the region embedding that summarizes region $r_n$'s embedding sequence. $p_{t, n}$ measures the importance of $\bm b_{\tau,n}$ to $\bm u_n$, where $\bm b_{\tau,n}$ is the embedding at time $\tau$. $\bm w_0 \in \mathbb{R}^{D}$ is a learnable parameter vector.

In our \name model, we propose to estimate the heterogeneity degree between two regions, to be reflective of their traffic distribution difference over time below:
\begin{equation}\label{eq:sa}\small
    q_{m,n} = \frac{\bm{u}_m^{\top} \bm{u}_n}{\Vert \bm{u}_m \Vert \Vert \bm{u}_n \Vert}.
\end{equation}
\noindent Note that a larger $q_{m,n}$ score indicates the higher traffic pattern dependencies between region $r_m$ and $r_n$, thus resulting in the lower heterogeneity degree.

% Given two region embeddings $\bm u_m$ and $\bm u_n$, we can measure their similarity as follows
% \begin{equation}\label{eq:sa}\small
%     q_{mn} = \frac{\bm{u}_m^{\top} \bm{u}_n}{\Vert \bm{u}_m \Vert \Vert \bm{u}_n \Vert},
% \end{equation}
% where a larger $q_{mn}$ means a higher similarity among the two regions. Regions with higher similarity have lower heterogeneity in traffic flow distribution, and vice versa.

\subsubsection{Heterogeneity-guided Data Augmentation.}
%\subsubsection{Data Augmentation Guided by Heterogeneity Measurement.} 

In our \name, we propose to perform data augmentation from both the traffic-level and graph topology-level elaborated below:\\\vspace{-0.12in}

% We use $p$ and $q$ in Eq. \eqref{eq:ta} and \eqref{eq:sa} to augment the input $\mathcal{G} = \left(\mathcal{V}, \mathcal{E}, \bm A, \mathcal{X}_{t-T:t}\right)$ in terms of region-attribute-level augmentation and graph-topology-level augmentation, respectively.

\noindent \textbf{Traffic-level Augmentation}. Inspired by the data augmentation strategy in~\cite{zhu2021graph}, we design an augmentation operator over the constructed traffic tensor $\mathcal{X}_{t-T:t}$, which is adaptive to the learned time-aware traffic pattern dependencies of each region. In particular, we aim to mask less relevant traffic volume at $\tau$-th time step of region $r_n$ against noise perturbation, based on a derived mask probability ${\rho}_{\tau,n}$ draw from a Bernoulli distribution \ie ${\rho}_{\tau,n} \sim \mathrm{Bern}(1-p_{\tau,n})$. The higher ${\rho}_{\tau,n}$ value indicates that region $r_n$'s traffic volume $\bm x_{\tau, n}$ at $\tau$-th time step is more likely to be masked, due to its lower relevance to the overall traffic regularities of region $r_n$. The augmented data with the traffic-level augmentation is denoted as $\tilde{\mathcal{X}}_{t-T:t}$.\\\vspace{-0.12in}

% In the region-attribute-level augmentation, we augment $\mathcal{X}_{t-T:t}$. Since $p_{t,n}$ measures the importance of attributes at time $t$ of region $r_n$, we randomly mask less important attributes. Specifically, we draw from a Bernoulli distribution \ie ${\rho}_{t,n} \sim \mathrm{Bern}(1-p_{t,n})$. If $\rho_{t,n}$ is equal to $1$, we mask the region flow $\bm x_{t,n}\in \mathbb{R}^{2}$ as a zero vector. Otherwise, we keep the region flow unchanged. By repeating this process over all time steps in every region, we obtain the output of the region-attribute-level augmentation $\tilde{\mathcal{X}}_{t-T:t}$.

\noindent \textbf{Graph Topology-level Augmentation}.
In addition to the traffic-level augmentation, we propose to further perform the topology-level augmentation over the region traffic flow graph $\mathcal{G}$. By doing so, \name can not only debias the region connections with low inter-correlated traffic patterns, but also capture the long-range region dependencies with the global urban context. Towards this end, $i$) Given two spatially adjacent regions $r_m$ and $r_n$, their connection edge $(r_m, r_n) \in \mathcal{E}$ will be masked if they are not highly dependent in terms of their traffic regularities, measured by the high heterogeneity degree $q_{m,n}$. The mask probability ${\rho}_{m,n}$ is drawn from a Bernoulli distribution \ie ${\rho}_{m,n} \sim \mathrm{Bern}(1-q_{m,n})$. $ii$) Given two non-adjacent regions, the low heterogeneity degree $q_{m,n}$ will result in adding an edge between $r_m$ and $r_n$ based on the masking probability drawn from a Bernoulli distribution, $\mathrm{Bern}(q_{m,n})$ similarly.

% In the graph-topology-level augmentation, we augment $\mathcal{E}$ and its corresponding $\bm A$. For each edge $(r_m, r_n) \in \mathcal{E}$, we use $1-q_{mn}$ as a measurement of heterogeneous relations, where $q_{mn}$ is from Eq.~\eqref{eq:sa}. Then, we draw from a Bernoulli distribution, \ie ${\rho}_{mn} \sim \mathrm{Bern}(1-q_{mn})$, and randomly remove the edge connecting two heterogeneous regions. If ${\rho}_{mn}$ is equal to $1$, we remove the edge $(r_m, r_n)$, otherwise we keep it. Similarly, we add edges by drawing from a Bernoulli distribution $\mathrm{Bern}(q_{mn})$. Because we remove edges connecting heterogeneous regions and add edges connecting similar regions, there are fewer heterogeneous regions connected and more similar regions connected in the augmentation graph, as shown in \figureautorefname~\ref{fig:adaaug}. 

After two augmentation phases, we obtain the augmented TFG $\tilde{\mathcal{G}} = \left(\mathcal{V}, \tilde{\mathcal{E}}, \tilde{\bm A}, \tilde{\mathcal{X}}_{t-T:t}\right)$, with the debiased traffic volume input $\tilde{\mathcal{X}}_{t-T:t}$ (traffic-level augmentation) and structure denoising $\tilde{\mathcal{E}}, \tilde{\bm A}$ (graph topology-level augmentation).

% Finally, we obtain the augmented TFG data, $\tilde{\mathcal{G}} = \left(\mathcal{V}, \tilde{\mathcal{E}}, \tilde{\bm A}, \tilde{\mathcal{X}}_{t-T:t}\right)$, where $\tilde{\mathcal{X}}_{t-T:t}$ is the output of the region-attribute-level augmentation, and $\tilde{\mathcal{E}}, \tilde{\bm A}$ are outputs of the graph-topology-level augmentation. Moreover, we name $\tilde{\mathcal{G}}$ the heterogeneity-eliminated TFG.

% \begin{figure}[t]
%     \centering
%     \subfigure[Before]{
%         \includegraphics[width=0.3\columnwidth]{figures/adaaug-a.pdf}
%     }\quad\quad
%     \subfigure[After]{
%         \includegraphics[width=0.3\columnwidth]{figures/adaaug-b.pdf}
%     }
%     \vspace{-0.2cm}\caption{Graph topology before and after the AdaAug. The different shapes represent different types of spatial regions.}\label{fig:adaaug}\vspace{-0.3cm}
% \end{figure}

\subsection{SSL for Spatial Heterogeneity Modeling}

% The objective of this module is to model the spatial heterogeneity. The spatial heterogeneity poses challenges when adjacent regions have heterogeneous traffic flow distributions. To address it, we design a self-supervised learning task called cluster assignment prediction. Since the heterogeneity-eliminated TFG removes relations among heterogeneous regions and build relations for similar regions, we exploit it to produce region cluster assignments as pseudo labels. Then, we predict them using the embeddings generated from the original input TFG, so as to encourage the model to be aware of spatial heterogeneity and to involve more non-heterogeneous relations when generating embeddings.

Given the heterogeneity-aware augmented TFG, we aim to enable the region embeddings to effectively preserve the spatial heterogeneity with auxiliary self-supervised signals.

To achieve this goal, we design a soft clustering-based self-supervised learning (SSL) task over regions, to map them into multiple latent representation spaces corresponding to diverse urban region functionalities (\eg residential zone, shopping mall, transportation hub). Specifically, we generate $K$ cluster embeddings $\{\bm c_1, \dots, \bm c_K\}$ (indexed by $k$) as latent factors for region clustering. Formally, the clustering process is performed with $\tilde{z}_{n,k} = \bm c_k^\top \tilde{\bm h}_n$. Here, $\tilde{\bm h}_n\in \mathbb{R}^{D}$ is the region embedding of region $r_n$ encoded from the augmented TFG $\tilde{\mathcal{G}}$. $\tilde{z}_{n,k}$ represents the estimated relevance score between region $r_n$'s embedding and the embedding $\bm c_k$ of the $k$-th cluster. Afterwards, the cluster assignment of region $r_n$ is generated with $\tilde{\bm z}_n = (\tilde{z}_{n,1}, \dots, \tilde{z}_{n,K})^\top$.

% We adopt a soft clustering method to generate cluster assignment for each region. Specifically, we have a set of cluster embeddings $\{\bm c_1, \dots, \bm c_K\}$, where $\bm c_k\in \mathbb{R}^{D}$ is the embedding of the $k$-th cluster. Given a region $r_n$, we define its augmented embedding as $\tilde{\bm h}_n\in \mathbb{R}^{D}$, which is generated by the ST encoder from $\tilde{\mathcal{G}}$. Then, we cluster the region by
% \begin{equation}\small
%     \tilde{z}_{nk} = \bm c_k^\top \tilde{\bm h}_n,
%\end{equation}
% where $\tilde{z}_{nk}$ represents the measurement of region $r_n$ belonging to the $k$-th cluster. The cluster assignment of region $r_n$ is denoted by $\tilde{\bm z}_n = (\tilde{z}_{n1}, \dots, \tilde{z}_{nK})^\top$. 

To provide self-supervised signals based on the heterogeneity-aware soft clustering paradigm for augmentation, the auxiliary learning task is designed to predict the cluster assignment using the region embedding ${\bm h_n}$ encoded from the original TFG $\mathcal{G}$ as: $\hat{z}_{n,k} = \bm c_k^{\top}{\bm h_n}$, where $\hat{\bm z}_{n,k}$ is the predicted assignment score for $\tilde{\bm z}_{n,k}$. The self-supervised augmented task is optimized as follows:
\begin{equation}\label{eq:xent}\small
    \ell({\bm h}_n, \tilde{\bm z}_n) = -\sum_k \tilde{z}_{n,k} \log\frac{\exp \left(\hat{z}_{n,k}/\gamma \right)}{\sum_{j} \exp \left(\hat{z}_{n,j}/\gamma \right)},
\end{equation}
\noindent where $\gamma$ is the temperature parameter to control the smoothing degree of softmax output. The overall self-supervised objective over all regions is defined as follows:
\begin{equation}\label{eq:loss_s}\small
    \mathcal{L}_s =\sum_{n=1}^{N} \ell({\bm h}_n, \tilde{\bm z}_n).
\end{equation}

% Then, we predict each assignment item using ${\bm h_n}$, \ie 
% \begin{equation}
%     \hat{z}_{nk} = \bm c_k^{\top}{\bm h_n},
% \end{equation}
% where $\hat{\bm z}_{nk}$ is the prediction of assignment $\tilde{\bm z}_{nk}$. The loss is defined as
% \begin{equation}\label{eq:xent}\small
%     \ell({\bm h}_n, \tilde{\bm z}_n) = \sum_k \tilde{z}_{nk} \log\frac{\exp \left(\hat{z}_{nk}/\tau \right)}{\sum_{j} \exp \left(\hat{z}_{nj}/\tau \right)},
% \end{equation}
% where $\tau$ is an adjustable temperature parameter. By adding the loss of every region, we have the final loss of the cluster assignment prediction task as follows
% \begin{equation}\label{eq:loss_s}\small
%     \mathcal{L}_s =\sum_{n=1}^{N} \ell({\bm h}_n, \tilde{\bm z}_n).
% \end{equation}

By incorporating the supervision on ${\bm h}_n$ with the heterogeneity-aware region cluster assignment $\tilde{\bm z}_n$, we make the region embedding ${\bm h}_n$ to be reflective of spatial heterogeneity within the global urban space.\\\vspace{-0.12in}

% The embedding ${\bm h}_n$ involves heterogeneous relations that exist in $\mathcal{G}$, while $\tilde{\bm z}_n$ is produced by using the heterogeneity-eliminated TFG $\tilde{\mathcal{G}}$. Therefore, using $\tilde{\bm z}_n$ to supervise ${\bm h}_n$ can encourage ${\bm h}_n$ to involve more non-heterogeneous relations.

\noindent \textbf{Distribution Regularization for Region Clustering}.
In our heterogeneity-aware region clustering paradigm, we generate the cluster assignment matrix $\tilde{\bm Z} = (\tilde{\bm z}_1, \dots, \tilde{\bm z}_N)^\top \in \mathbb{R}^{N \times K}$ as self-supervised signals for generative data augmentation. However, two issues need to be addressed to fit the true distribution of regional characteristics in urban space: $i$) Since $\tilde{\bm Z}$ is produced by matrix production, there is no guarantee that each region's cluster assignment sums up to 1, \ie $\tilde{\bm Z} \bm 1_K = \bm 1_N$, where $\bm 1_N$ denotes an $N$-dimensional vector of all ones. $ii$) To avoid the trivial solution that every region has the same assignment, we employ the principle of maximum entropy, \ie $\tilde{\bm Z}^\top \bm 1_N = \frac{N}{K} \bm 1_K$. This encourages all regions to be equally partitioned by the clusters. To tackle these two issues, we define a feasible solution set as:
\begin{equation}\small
    \tilde{\mathcal{Z}} = \left\{\tilde{\bm Z} \in \mathbb{R}_{+}^{N \times K} \middle| \tilde{\bm Z} \bm 1_K = \bm 1_N, \tilde{\bm Z}^\top \bm 1_N = \frac{N}{K} \bm 1_K \right\}.
\end{equation}

% In Eq.~\eqref{eq:loss_s}, we use $\tilde{\bm z}_n$ as supervision signals. However, directly using it may lead to two issues. To this end, we pre-process $\tilde{\bm z}_n$ before using it. Specifically, given the cluster assignment matrix $\tilde{\bm Z} = (\tilde{\bm z}_1, \dots, \tilde{\bm z}_N)^\top \in \mathbb{R}^{N \times K}$, the two issues are given as follows: $i$) Since $\tilde{\bm Z}$ is produced by matrix production, there is no guarantee that each region's cluster assignment sums up to 1, \ie $\tilde{\bm Z} \bm 1_K = \bm 1_N$, where $\bm 1_N$ denotes an $N$-dimensional vector of all ones. $ii$) To avoid the trivial solution that every region has the same assignment, we employ the principle of maximum entropy, \ie $\tilde{\bm Z}^\top \bm 1_N = \frac{N}{K} \bm 1_K$. This encourages all regions are equally partitioned by the clusters. To address the two issues, we define a feasible solution set as 
% \begin{equation}\small
%     \tilde{\mathcal{Z}} = \left\{\tilde{\bm Z} \in \mathbb{R}_{+}^{N \times K} \middle| \tilde{\bm Z} \bm 1_K = \bm 1_N, \tilde{\bm Z}^\top \bm 1_N = \frac{N}{K} \bm 1_K \right\}.
% \end{equation}

For any assignment $\tilde{\bm Z} \in \tilde{\mathcal{Z}}$, we can use it to map the embedding matrix $\tilde{\bm H} = (\tilde{\bm h}_1, \dots, \tilde{\bm h}_N)^\top \in \mathbb{R}^{N \times D}$ into the cluster matrix $\bm C = (\bm c_1, \dots, \bm c_K)^\top \in \mathbb{R}^{K \times D}$. Thus, we search for the optimal solution by maximizing the similarity between the embeddings and the clusters, \ie 
\begin{equation}\small
    \max_{\tilde{\bm Z} \in \mathcal{Z}} \mathrm{tr}\left(\tilde{\bm Z} \bm C \tilde{\bm H}^\top\right) + \epsilon H(\tilde{\bm Z}), 
\end{equation}
where $\mathrm{tr}(\cdot)$ is the trace operator that sums elements on the main diagonal of a square matrix, $H(\tilde{\bm Z})$ is the entropy function defined as $- \sum_{n,k} \tilde{z}_{n,k}\log \tilde{z}_{n,k}$, and $\epsilon$ is a parameter that controls the smoothness of the assignment. Finally, the original assignment in Eq. \eqref{eq:loss_s} is replaced with the optimal solution. Refer to the Appendix for the solution procedure.

\subsection{SSL for Temporal Heterogeneity Modeling}

% \eg the traffic in the morning peak shows an increasing trend compared with the nighttime

In this component, we further design a self-supervised learning (SSL) task to inject the temporal heterogeneity into time-aware region embeddings, by enforcing the divergence among time step-specific traffic pattern representations. 
% This module aims to capture the temporal heterogeneity. The temporal heterogeneity means traffic trends varying with different time periods. However, in the same time period, the trends in most regions may be consistent. To exploit it, we introduce a self-supervised learning task called traffic trend classification. It first generates a city-level embedding from region-level embeddings. Then, it distinguishes the two types of embeddings in the same time period from those in different periods.
% Specifically, given a time period $(t-T, t)$ and a region $r_n$, we denote the region embeddings from $\mathcal{G}$ and $\tilde{\mathcal{G}}$ as $\bm h_{t, n}$ and $\tilde{\bm h}_{t, n}$. Then, we fuse them into an ensemble embedding as follows
% \begin{equation}\small
%     \bm{v}_{t, n} = \bm{w_1} \odot \bm{h}_{t, n} + \bm{w_2} \odot \tilde{\bm h}_{t, n},
% \end{equation}
% where $\odot$ is the element-wise product, and $\bm w_1, \bm w_2$ are learnable parameters. 

Specifically, we firstly fuse the encoded time-aware region embeddings from both the original and augmented TFGs:
\begin{equation}\small
    \bm{v}_{t, n} = \bm{w_1} \odot \bm{h}_{t, n} + \bm{w_2} \odot \tilde{\bm h}_{t, n},
\end{equation}
\noindent where $\odot$ is the element-wise product. $\bm w_1, \bm w_2$ are learnable parameters. After that, we generate the city-level representation $\bm s_{t}$ at the time step $t$ through aggregating embeddings of all regions ($\sigma$ is the sigmoid function):
\begin{equation}\small
    \bm s_{t} = \sigma \left(\frac{1}{N} \sum_{n=1}^{N} \bm v_{t,n} \right).
\end{equation}

% To obtain the city-level embedding, we summarize region embeddings of all regions by
% \begin{equation}\small
%     \bm s_{t} = \sigma \left(\frac{1}{N} \sum_{n=1}^{N} \bm v_{t,n} \right),
% \end{equation}
% where $\bm s_{t} \in \mathbb{R}^{D}$ is the city-level summary, and $\sigma$ is the sigmoid function.

To enhance the representation discrimination ability among different time steps, we treat the region-level and city-level embeddings ($\bm v_{t, n}, \bm s_{t}$) from the same time step as the positive pairs in our SSL task, and the embeddings from different time steps as negative pairs. With this design, the auxiliary supervision of positive pairs will encourage the consistency of time-specific citywide traffic trends (\eg rush hours, weather factors), while the negative pairs help in capturing the temporal heterogeneity across different time steps. Formally, the temporal heterogeneity-enhanced SSL task is optimized with the following loss with cross-entropy metric:
\begin{equation}\label{eq:loss_t}\small
    \mathcal{L}_t =-\left(\sum_{n=1}^{N} \log g\left(\bm v_{t, n}, \bm s_{t}\right) + \sum_{n=1}^{N} \log \left(1-g\left(\bm v_{t', n}, \bm s_{t}\right)\right)\right),
\end{equation}
\noindent where $t$ and $t'$ denote two different time steps. $g$ is a criterion function defined as $g\left(\bm v_{t, n}, \bm s_{t}\right) = \sigma\left(\bm v_{t,n}^\top \bm W_3 \bm s_{t} \right)$. $\bm W_3 \in \mathbb{R}^{N \times N}$ is the learnable transformation matrix.

% The city-level summary contains the temporal information of all regions. For example, during morning rush hours, most region's flow could be increase, allowing the city-level summary to describe this increasing trend. Therefore, we can encourage the model to distinguish whether a region's embedding satisfies a trend, and make the model aware of different time periods while making predictions. 

% Specifically, we use the embeddings $\left(\bm v_{t, n}, \bm s_{t}\right)$ of the same time period as a positive pair, and the embeddings of different time periods as a negative pair. To distinguish positive and negative pairs, we employ the cross entropy loss as
% \begin{equation}\label{eq:loss_t}\small
%     \mathcal{L}_t =\sum_{n=1}^{N} \log g\left(\bm v_{t, n}, \bm s_{t}\right) + \sum_{n=1}^{N} \log \left(1-g\left(\bm v_{t', n}, \bm s_{t}\right)\right),
% \end{equation}

% where $t'$ represents another time period. Here, $g$ is a criterion function defined as $g\left(\bm v_{t, n}, \bm s_{t}\right) = \sigma\left(\bm v_{t,n}^\top \bm W_3 \bm s_{t} \right)$, where $\bm W_3 \in \mathbb{R}^{N \times N}$ is learnable, and $\sigma$ is the sigmoid function.

\subsection{Model Training}

In the learning process of our \name, we feed the embedding $\bm h_n \in \bm H$ of each region $r_n$ into an MLP structure to enable the traffic flow prediction at the future time step $t+1$ as:
\begin{equation}
    \hat{\bm x}_{t+1, n} = \mathrm{MLP}(\bm h_n),
\end{equation}
\noindent where $\hat{\bm x}_{t+1, n}$ is the predicted result. The model is optimized by minimizing the loss function below:
\begin{equation}\label{eq:loss_p}\small
    \mathcal{L}_p = \sum_{n=1}^{N} \lambda \left|x_{t+1,n}^{(0)} - \hat{x}_{t+1,n}^{(0)}\right|+ (1-\lambda) \left|x_{t+1,n}^{(1)} - \hat{x}_{t+1,n}^{(1)}\right|,
\end{equation}
where $x_{t+1,n}^{(0)}, x_{t+1,n}^{(1)}$ denote the ground truth of inflow and outflow respectively. $\lambda$ is a parameter to balance the influence of each type of traffic flow.

% Given a region $r_n$, we take the region embedding $\bm h_n \in \bm H$ and feed it into a MLP to enable the traffic flow prediction at the future time step $t+1$ by 
% \begin{equation}
%     \hat{\bm x}_{t+1, n} = \mathrm{MLP}(\bm h_n),
% \end{equation}
% where $\hat{\bm x}_{t+1, n}$ is the prediction result. The prediction loss is defined as
% \begin{equation}\label{eq:loss_p}\small
%     \begin{split}
%         \mathcal{L}_p = \sum_{n=1}^{N} &\lambda |x_{t+1,n,0} - \hat{x}_{t+1,n, 0}|\\
%         &\quad+ (1-\lambda) |x_{t+1,n,1} - \hat{x}_{t+1,n,1}|,
%     \end{split}
% \end{equation}
% \begin{equation}\label{eq:loss_p}\small
%     \mathcal{L}_p = \sum_{n=1}^{N} \lambda \left|x_{t+1,n}^{(0)} - \hat{x}_{t+1,n}^{(0)}\right|+ (1-\lambda) \left|x_{t+1,n}^{(1)} - \hat{x}_{t+1,n}^{(1)}\right|,
% \end{equation}
% where $x_{t+1,n}^{(0)}, x_{t+1,n}^{(1)}$ denote the ground truth of inflow and outflow respectively. $\lambda$ is a parameter to balance the influence of each kind of flow.

Finally, we obtain the overall loss by incorporating the self-supervised spatial and temporal heterogeneity modeling losses in Eq. \eqref{eq:loss_s} and \eqref{eq:loss_t} into the joint learning objective:
\begin{equation}
    \mathcal{L}_{joint} = \mathcal{L}_p + \mathcal{L}_s + \mathcal{L}_t.
\end{equation}

Our model can be trained via the back-propagation algorithm. The entire training procedure can be summarized into four stages: $i$) given a TFG $\mathcal{G}$, we generate a region embedding matrix $\bm H$ by the ST encoder. $ii$) Meanwhile, we perform adaptive augmentation to refine $\mathcal{G}$ as $\tilde{\mathcal{G}}$, which is fed into the shared ST encoder to output $\tilde{\bm H}$. $iii$) By using $\bm H$ and $\tilde{\bm H}$, we calculate the losses $\mathcal{L}_s$, $\mathcal{L}_t$, and $\mathcal{L}_p$ that are used to produce the joint loss $\mathcal{L}_{joint}$. $iv$). We employ the back-propagation algorithm to train \name until $\mathcal{L}_{joint}$ converges.

\section{Experiments}\label{sec:expt}

In this section, we evaluate the performance of \name on a series of experiments over several real-world datasets, which are summarized to answer the following research questions:

\bitem{RQ1}: How is the overall traffic prediction performance of \name as compared to various baselines?

\bitem{RQ2}: How do designed different sub-modules contribute to the model performance?

\bitem{RQ3}: How does \name perform with regard to heterogeneous spatial regions and different time periods? 

\bitem{RQ4}: How do the augmented graph and learned representations benefit the model?

\subsection{Experimental Settings}

\subsubsection{Data Description.} We evaluate our model on two types of public real-world traffic datasets summarized in \tableautorefname~\ref{tab:dataset}.

The first kind is about bike rental records in New York City. {\bf NYCBike1}~\cite{zhang2017deep} spans from 04/01/2014 to 09/30/2014, and {\bf NYCBike2}~\cite{yao2019revisiting} spans from 07/01/2016 to 08/29/2016. They are all measured every 30 minutes. The second kind is about taxi GPS trajectories. {\bf NYCTaxi}~\cite{yao2019revisiting} spans from 01/01/2015 to 03/01/2015. Its time interval is half an hour. {\bf BJTaxi}~\cite{zhang2017deep}, collected in Beijing, spans from 03/01/2015 to 06/30/2015 on an hourly basis. 

For all datasets, previous 2-hour flows as well as previous 3-day flows around the predicted time are used to predict the flows for the next time step. We use a sliding window strategy to generate samples, and then split each dataset into the training, validation, and test sets with a ratio of 7:1:2.

\subsubsection{Evaluation Metrics \& Baselines.} In our experiments, two common metrics are used for evaluation: Mean Average Error (MAE) and Mean Average Percentage Error (MAPE). We compare our proposed \name with 8 baselines that fall into three categories. 
% $i$) \emph{Traditional time series prediction approaches:} We take ARIMA and support vector regression (SVR) as baselines. The history traffic flow-sequence are treated as purely time series to predict the future states without consideration of spatial information. $ii$) \emph{Spatial-temporal prediction methods:} The convolution-based ST-ResNet~\cite{zhang2017deep}, the graph convolution-based ST-GCN~\cite{yu2018spatio} and the attention-based GMAN~\cite{zheng2020gman} are used for comparison. $iii$) \emph{Spatial-temporal methods considering heterogeneity:} We use state-of-the-art methods that consider traffic data heterogeneity such as AGCRN~\cite{bai2020adaptive}, STSGCN~\cite{song2020spatial} and STFGNN~\cite{li2021spatial} for comparison.

{\bf \noindent Traditional Time Series Prediction Approaches:}\\
\bitem{ARIMA}~\cite{kumar2015short}: it is a
classical time series prediction model.\\
\bitem{SVR}~\cite{castro2009online}: it is a regression model widely used for time series analysis.\\\vspace{-0.12in}

{\bf \noindent Spatial-Temporal Traffic Prediction Methods:}\\
\bitem{ST-ResNet}~\cite{zhang2017deep}: it is a convolution-based model that constructs multiple traffic time series to capture the temporal dependencies and utilizes residual convolution to model the spatial correlations.\\
\bitem{STGCN}~\cite{yu2018spatio}: it is a graph convolution-based model that combines 1D convolution to capture spatial and temporal correlations, respectively.\\
\bitem{GMAN}~\cite{zheng2020gman}: it is an attention-based predictive model that adopts an encoder-decoder architecture.\\\vspace{-0.12in}

{\bf \noindent Spatial-Temporal Methods Considering Heterogeneity:}\\
\bitem{AGCRN}~\cite{bai2020adaptive}: it enhances the traditional graph convolution by adaptive modules and combines them into recurrent networks to capture spatial-temporal correlations.\\
\bitem{STSGCN}~\cite{song2020spatial}: it captures the complex localized spatial-temporal correlations through a spatial-temporal synchronous modeling mechanism.\\
\bitem{STFGNN}~\cite{li2021spatial}: it integrates with STFGN module and a novel gated CNN module, and captures hidden spatial dependencies by a data-driven graph and its further fusion with given spatial graphs.

Methods in the last category model the traffic heterogeneity by using multiple parameter spaces.

\begin{table}[t]\small
    \centering
    \setlength{\tabcolsep}{1.5mm}
    \begin{tabular}{rcccc}
      \toprule 
      {Data type} & \multicolumn{2}{c}{Bike rental} & \multicolumn{2}{c}{Taxi GPS} \\
      \midrule
      Dataset & NYCBike1 & NYCBike2 & NYCTaxi & BJTaxi \\
      Time interval & 1 hour & 30 min & 30 min & 30 min \\
      \# regions & 16$\times$8 & 10$\times$20 & 10$\times$20 & 32$\times$32 \\
      \# taxis/bikes & 6.8k+ & 2.6m+ & 22m+ & 34k+ \\
      \bottomrule
    \end{tabular}%
    % \vspace{-.2cm}
    \caption{Statistics of Datasets.}\vspace{-.3cm}
    \label{tab:dataset}%
  \end{table}%

\begin{figure}[t]
    \centering
    \subfigure[ST-ResNet]{
        \includegraphics[width=0.3\columnwidth]{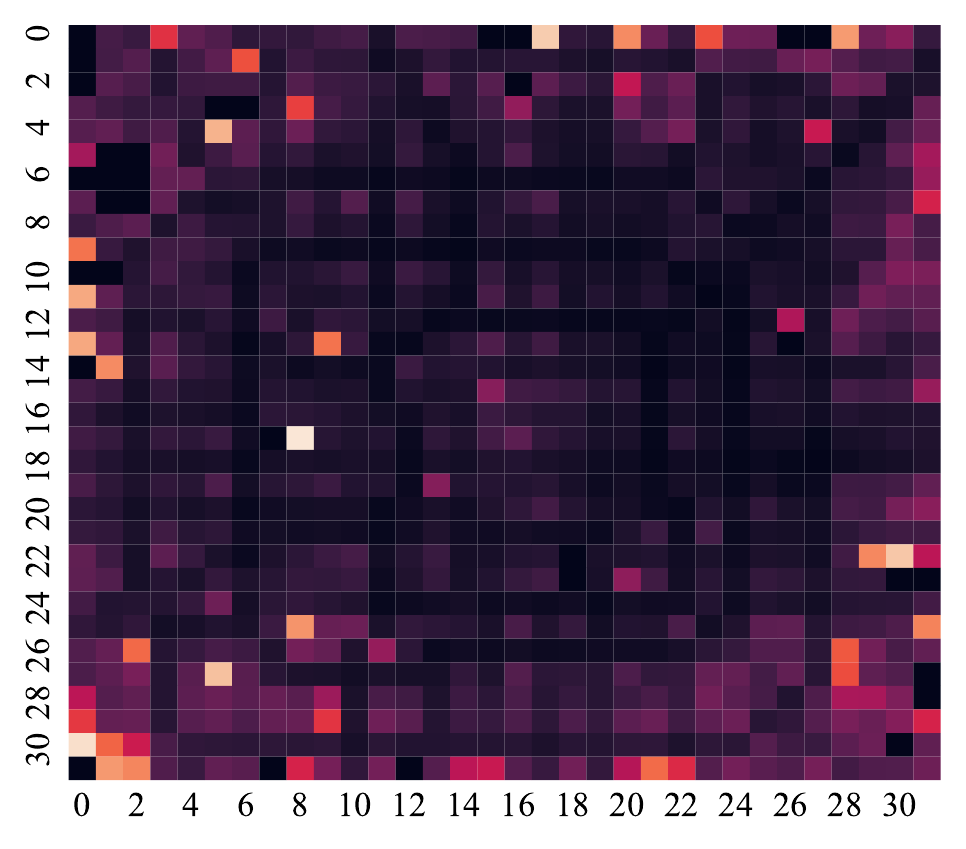}
    }
    \subfigure[AGCRN]{
        \includegraphics[width=0.3\columnwidth]{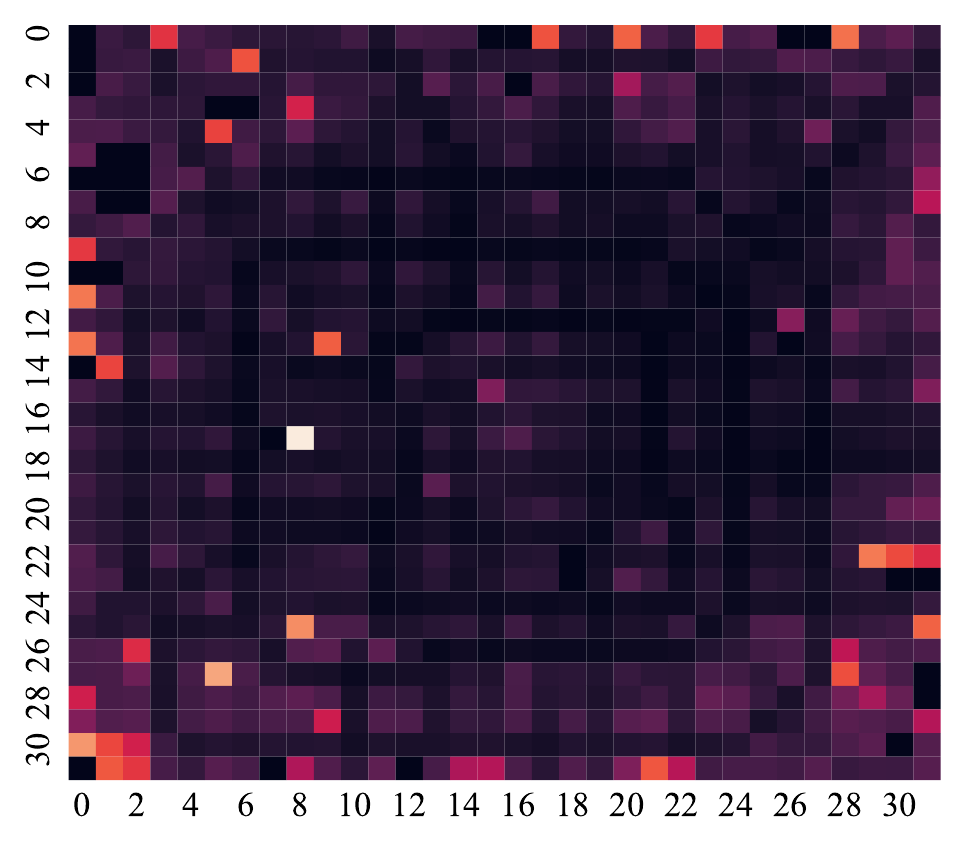}
    }
    \subfigure[\name]{
        \includegraphics[width=0.31\columnwidth]{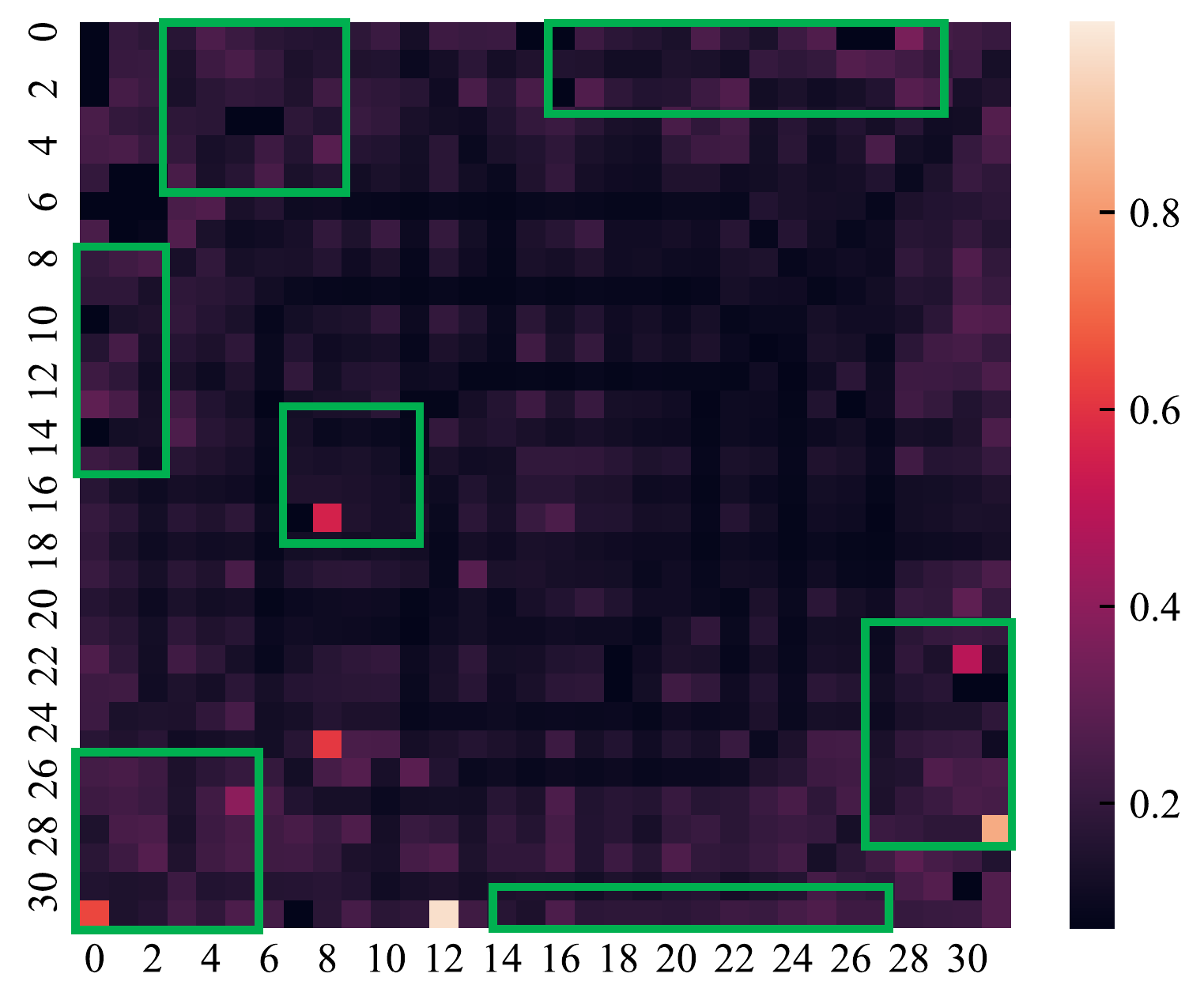}
    }
    \vspace{-.3cm}
    \caption{Visualization of traffic prediction errors.}\label{fig:pred_error}\vspace{-.4cm}
\end{figure}

\begin{table*}[t]\small
    \centering
    \setlength{\tabcolsep}{1mm}
    \renewcommand{\arraystretch}{1.1}
      \begin{tabular}{c||cc|cc|ccc|ccc|c}
      \toprule
        Dataset    & \multicolumn{1}{c|}{Metric} & Type  & ARIMA & SVR   & ST-ResNet & STGCN & GMAN  & AGCRN & STSGCN & STFGNN & \name \\
      \hline
      \midrule
    \multirow{4}{*}{\rotatebox{0}{NYCBike1}} & \multicolumn{1}{c|}{\multirow{2}{*}{MAE}} & In    & 10.66  & 7.27  & 5.53±0.06 & 5.33±0.02 & 6.77±3.42 & 5.17±0.03 & 5.81±0.04 & 6.53±0.10  & \textbf{4.94±0.02} \\
        & \multicolumn{1}{c|}{} & Out   & 11.33  & 7.98  & 5.74±0.07 & 5.59±0.03 & 7.17±3.61 & 5.47±0.03 & 6.10±0.04 & 6.79±0.08 & \textbf{5.26±0.02} \\
        \cline{2-12}        & \multicolumn{1}{c|}{\multirow{2}{*}{MAPE}} & In    & 33.05  & 25.39  & 25.46±0.20 & 26.92±0.08 & 31.72±12.29 & 25.59±0.22 & 26.51±0.32 & 32.14±0.23 & \textbf{23.69±0.11} \\
        & \multicolumn{1}{c|}{} & Out   & 35.03  & 27.42  & 26.36±0.50 & 27.69±0.14 & 34.74±17.04 & 26.63±0.30 & 27.56±0.39 & 32.88±0.19 & \textbf{24.60±0.27} \\
%   \cline{2-12}          & \multicolumn{2}{c|}{\# params} & -     & -     & 1790k & 75k   & 212k  & 750k  & ?     & ?     & 196k \\
      \midrule
      \midrule
      \multirow{4}{*}{\rotatebox{0}{NYCBike2}} & \multicolumn{1}{c|}{\multirow{2}{*}{MAE}} & In    & 8.91  & 12.82  & 5.63±0.14 & 5.21±0.02 & 5.24±0.13 & 5.18±0.03 & 5.25±0.03 & 5.80±0.10 & \textbf{5.04±0.03} \\
          & \multicolumn{1}{c|}{} & Out   & 8.70  & 11.48  & 5.26±0.08 & 4.92±0.02 & 4.97±0.14 & 4.79±0.04 & 4.94±0.05 & 5.51±0.11  & \textbf{4.71±0.02} \\
          \cline{2-12}         & \multicolumn{1}{c|}{\multirow{2}{*}{MAPE}} & In    & 28.86  & 46.52  & 32.17±0.85 & 27.73±0.16 & 27.38±1.13 & 27.14±0.14 & 29.26±0.13 & 30.73±0.49 & \textbf{22.54±0.10} \\
          & \multicolumn{1}{c|}{} & Out   & 28.22  & 41.91  & 30.48±0.86 & 26.83±0.21 & 26.75±1.14 & 26.17±0.22 & 28.02±0.23 & 29.98±0.46  & \textbf{21.17±0.13} \\
% \cline{2-12}          & \multicolumn{2}{c|}{\# params} & -     & -     & 1790k & 75k   & 212k  & 750k  & ?     & ?     & 196k \\
\midrule
\midrule
    \multirow{4}{*}{\rotatebox{0}{NYCTaxi}} & \multicolumn{1}{c|}{\multirow{2}{*}{MAE}} & In    & 20.86  & 52.16  & 13.48±0.14 & 13.12±0.04 & 15.09±0.61 & 12.13±0.11 & 13.69±0.11 & 16.25±0.38 & \textbf{11.99±0.12} \\
          & \multicolumn{1}{c|}{} & Out   & 16.80  & 41.71  & 10.78±0.25 & 10.35±0.03 & 12.06±0.39 & 9.87±0.04 & 10.75±0.17 & 12.47±0.25 & \textbf{9.78±0.09} \\
          \cline{2-12}          & \multicolumn{1}{c|}{\multirow{2}{*}{MAPE}} & In    & 21.49  & 65.10  & 24.83±0.55 & 21.01±0.18 & 22.73±1.20 & 18.78±0.04 & 22.91±0.44 & 24.01±0.30 & \textbf{16.38±0.10} \\
          & \multicolumn{1}{c|}{} & Out   & 21.23  & 64.06  & 24.42±0.52 & 20.78±0.16 & 21.97±0.86 & 18.41±0.21 & 22.37±0.16 & 23.28±0.47 & \textbf{16.86±0.23} \\
% \cline{2-12}          & \multicolumn{2}{c|}{\# params} & -     & -     & 1790k & 75k   & 212k  & 750k  & ?     & ?     & 196k \\
\midrule
\midrule
    \multirow{4}{*}{\rotatebox{0}{BJTaxi}} & \multicolumn{1}{c|}{\multirow{2}{*}{MAE}} & In    & 21.48  & 52.77  & 12.12±0.11 & 12.34±0.09 & 13.13±0.43 & 12.30±0.06 & 12.72±0.03 & 13.83±0.04 & \textbf{11.31±0.03} \\
          & \multicolumn{1}{c|}{} & Out   & 21.60  & 52.74  & 12.16±0.12 & 12.41±0.08 & 13.20±0.43 & 12.38±0.06 & 12.79±0.03 & 13.89±0.04 & \textbf{11.40±0.02} \\
\cline{2-12}          & \multicolumn{1}{c|}{\multirow{2}{*}{MAPE}} & In    & 23.12  & 65.51  & 15.50±0.26 & 16.66±0.21 & 18.67±0.99 & 15.61±0.15 & 17.22±0.17 & 19.29±0.07 & \textbf{15.03±0.13} \\
          & \multicolumn{1}{c|}{} & Out   & 20.67  & 65.51  & 15.57±0.26 & 16.76±0.22 & 18.84±1.04 & 15.75±0.15 & 17.35±0.17 & 19.41±0.07 & \textbf{15.19±0.15} \\
% \cline{2-12}          & \multicolumn{2}{c|}{\# params} & -     & -     & 1790k & 75k   & 212k  & 750k  & ?     & ?     & 196k \\
    \bottomrule
    \end{tabular}%
    \vspace{-.2cm}
    \caption{Model comparison on four datasets in terms of MAE and MAPE (\%). In and Out represent the inflow and outflow.}
    \vspace{-.2cm}
    \label{tab:overall}%
  \end{table*}%

\subsubsection{Parameter Settings.} The \name is implemented with PyTorch. The embedding dimension $D$ is set as 64. Both the temporal and spatial convolution kernel sizes of ST encoder are set to 3. The perturbation ratios for both traffic-level and topology-level augmentations are set as 0.1. The training phase is performed using the Adam optimizer and the batch size of 32. The experiments of baseline evaluation are conducted with their released codes on the LibCity~\cite{wang2021libcity} platform. 
% More implementation details of our \name and other baselines are given in Appendix. 

\subsection{Performance Comparison (RQ1)}

\tablename{~\ref{tab:overall}} shows the comparison results of all methods. We run all deep learning models with 5 different seeds and report the average performance and their standard deviations. 

\subsubsection{Performance Superiority of \name.} According to Student's $t$-test at level 0.01, our \name significantly outperforms other competing baselines with regard to both metrics over all datasets. This demonstrates the effectiveness of \name in jointly modeling the spatial and temporal heterogeneity in a self-supervised manner. \figureautorefname{~\ref{fig:pred_error}} visualizes the prediction error ($|\hat{x}_n - x_n| / x_n$) of \name and two best performed baselines on BJTaxi dataset, where a brighter pixel means a larger error. The superiority of our model can still be observed, which is consistent with the quantitative results in \tablename{~\ref{tab:overall}}. Interestingly, \name exhibits a significant improvement in the suburban areas (green boxes in \figureautorefname{~\ref{fig:pred_error}}), which justifies the effectiveness of spatial heterogeneity modeling that transfers information among global similar regions.

\subsubsection{Performance Comparison between Baselines.} Spatio-temporal prediction methods outperform time series approaches in most cases, which suggests the necessity to capture spatial dependencies. The methods that take into account the heterogeneity of traffic data usually perform better than those that use shared parameters across different regions and time periods, indicating the rationality of learning spatial and temporal heterogeneity in traffic prediction.

\begin{figure}[t]
    \centering
    \includegraphics[width=0.98\columnwidth]{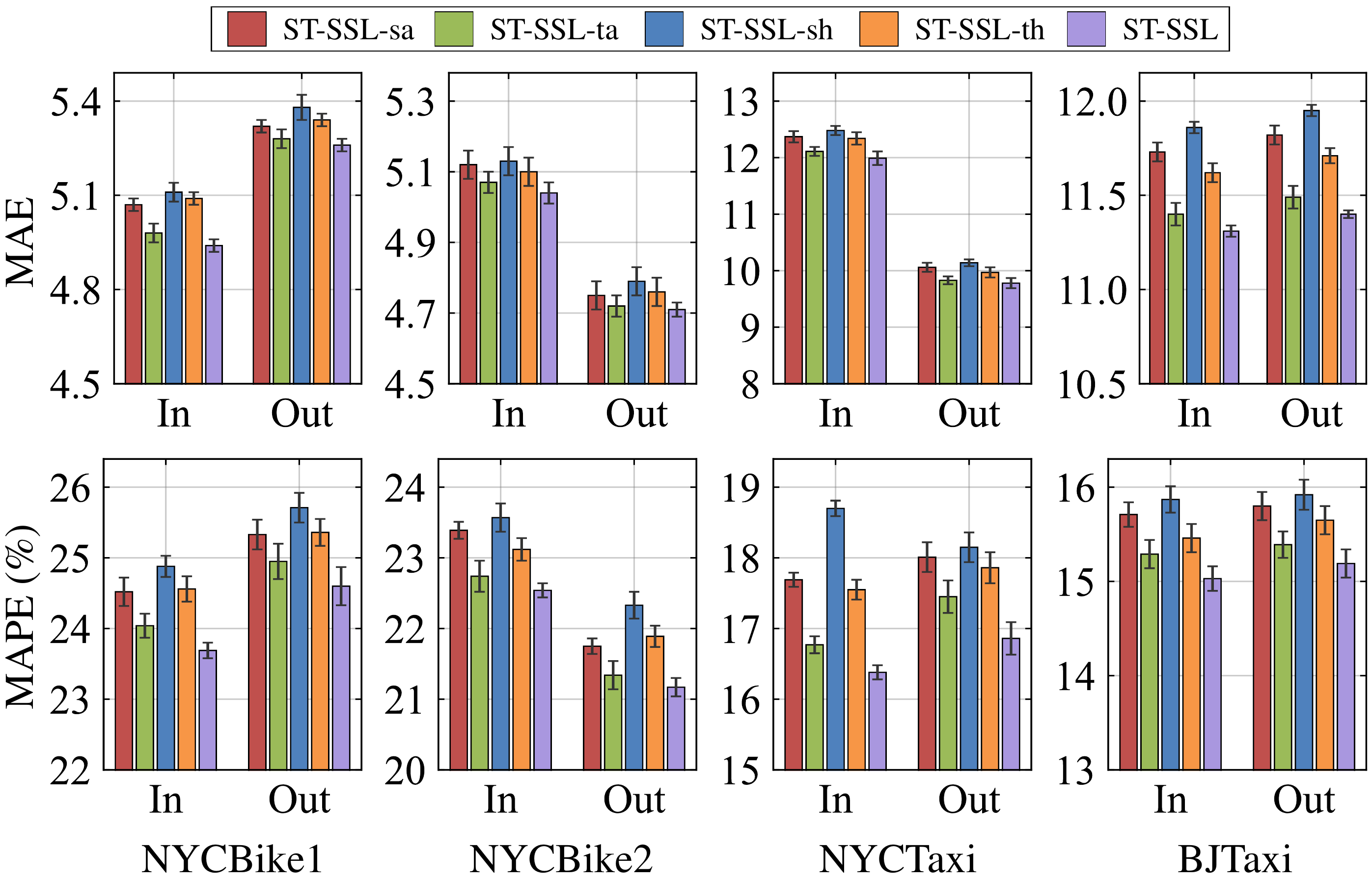}
    % \vspace{-.2cm}
    \caption{Ablation study of our proposed \name.}
    \vspace{-.3cm}
    \label{fig:ab}
\end{figure}

\subsection{Ablation Study (RQ2)}

To analyze the effects of sub-modules in our \name framework, we perform ablation studies with five variants:

\bitem{\name-sa}: This variant replaces heterogeneity-guided \underline{s}tructure \underline{a}ugmentation on graph topology with random edge removal and addition augmentations. 

\bitem{\name-ta}: This variant replaces heterogeneity-guided \underline{t}raffic-level \underline{a}ugmentation with random traffic volume masking augmentations.

\bitem{\name-sh}: This variant which disables \underline{s}patial \underline{h}eterogeneity modeling in the joint framework.
% to enable region embeddings to preserve spatial heterogeneity.

\bitem{\name-th}: This variant which disables \underline{t}emporal \underline{h}eterogeneity modeling in the joint framework.
% to inject the temporal heterogeneity into the time-aware region embeddings.

The results are presented in \figureautorefname{~\ref{fig:ab}}. We can observe that \name beats the variants with random augmentation, indicating the effectiveness of our adaptive heterogeneity-guided data augmentation at both traffic-level and graph structure-level. Moreover, \name consistently outperforms \name-sh and \name-th, which justifies the necessity to jointly model the spatial and temporal heterogeneity. In summary, each designed sub-module has a positive effect on performance improvement.

\begin{figure}[t]
    \centering
    \subfigure[Spatial clusters]{
        \includegraphics[width=0.44\columnwidth]{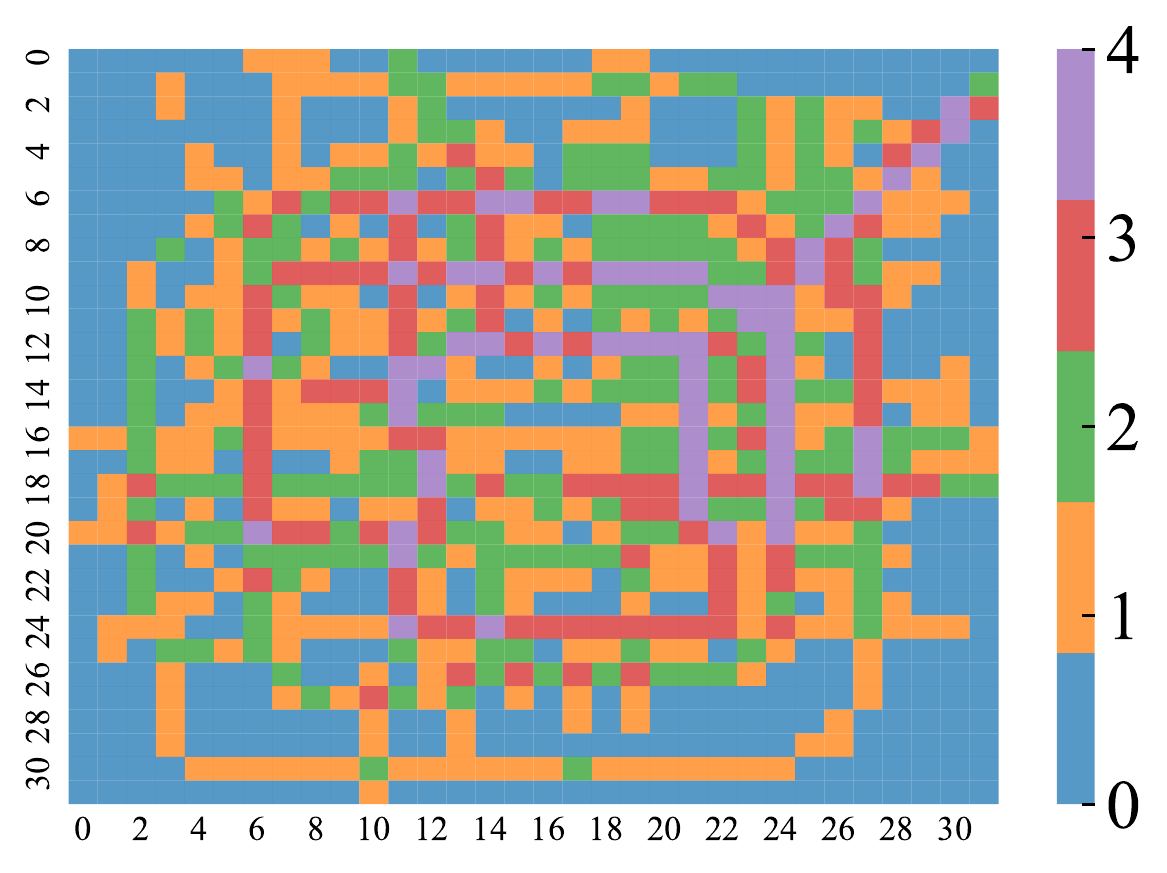}\label{fig:rb_sc}
    }\quad
    \subfigure[Spatial performance]{
        \includegraphics[width=0.47\columnwidth]{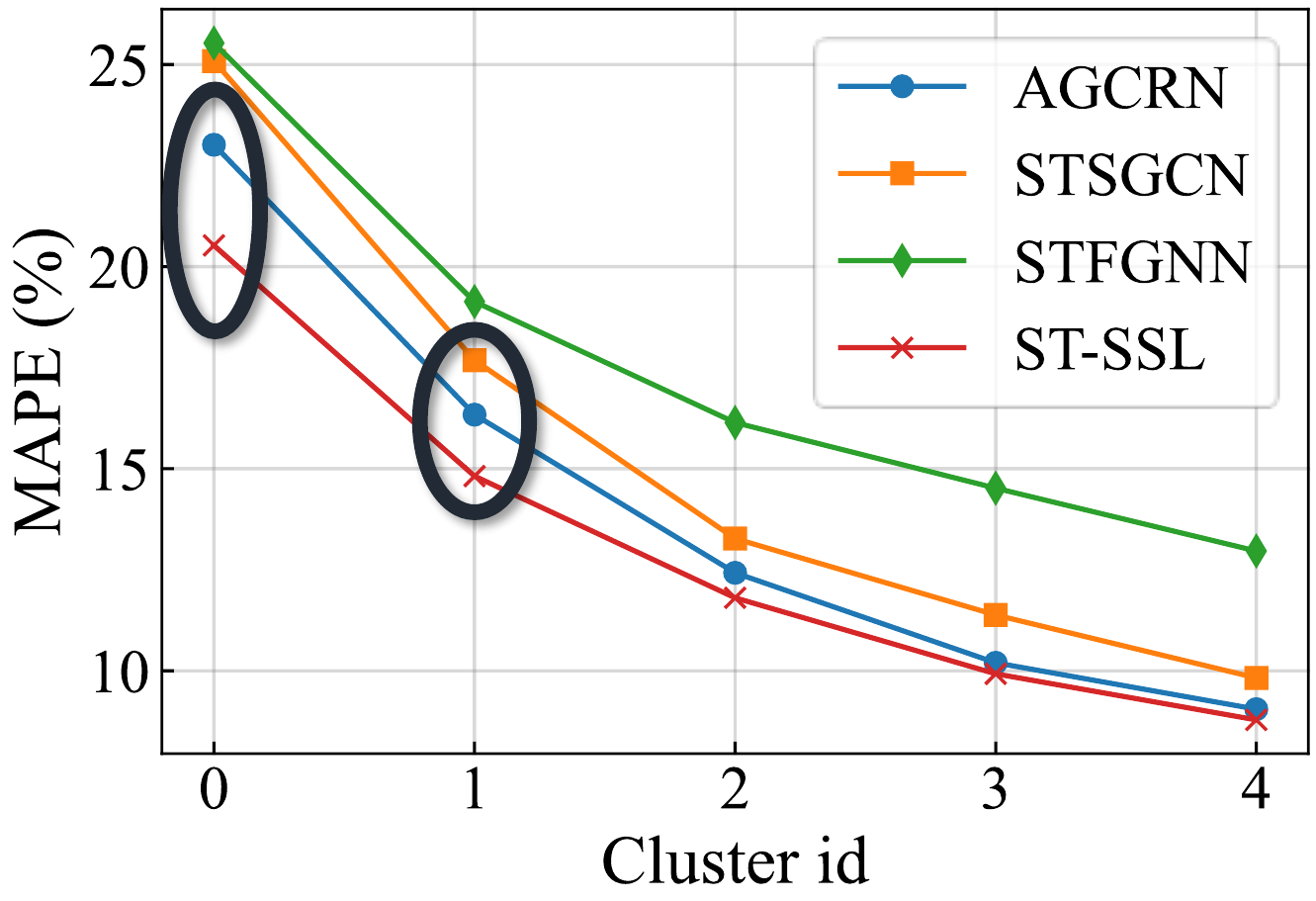}\label{fig:rb_sr}
    }
    \subfigure[Temporal categories]{
        \includegraphics[width=0.42\columnwidth]{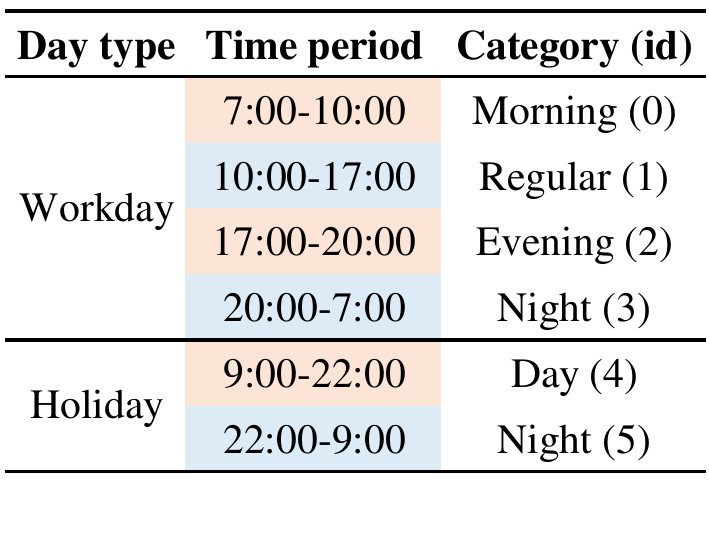}\label{fig:rb_tc}
    }\quad
    \subfigure[Temporal performance]{
        \includegraphics[width=0.47\columnwidth]{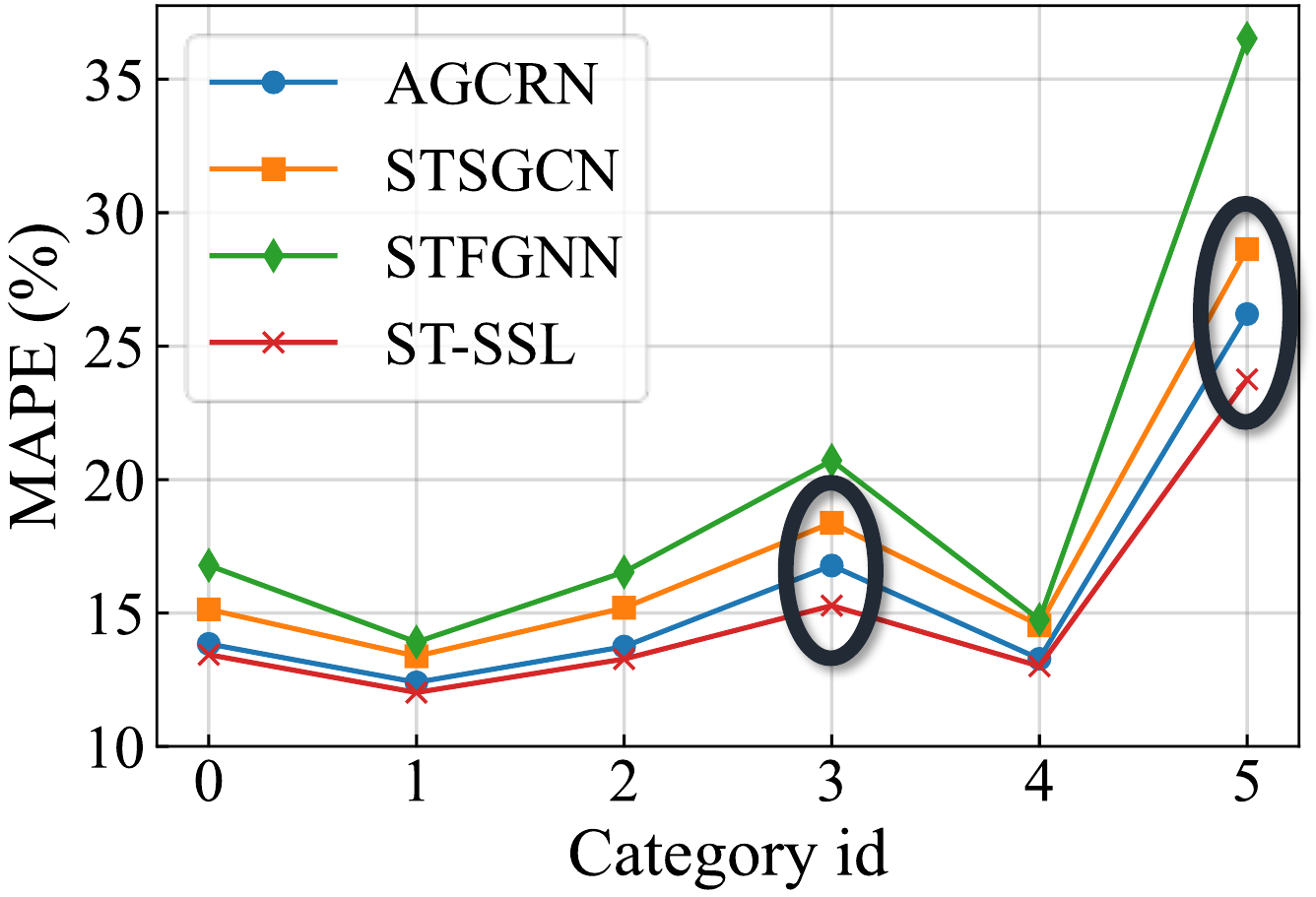}\label{fig:rb_tr}
    }
    \vspace{-.2cm}
    \caption{Prediction performance with regard to heterogeneous spatial regions and different time periods.}
    \vspace{-.3cm}
    \label{fig:robust}
\end{figure}

\subsection{Robustness Analysis (RQ3)}

To explore the robustness of our \name, we perform traffic prediction for spatial regions with heterogeneous data distributions and time periods with different patterns on BJTaxi. Specifically, we cluster regions by using traffic data statistics, \ie $(mean, median, standard~deviation)$ of their historical traffic flow. As shown in \figureautorefname{~\ref{fig:rb_sc}}, regions with smaller cluster id (next to the color bar) are usually located in suburbs that are less popular and thus have lower traffic. \figureautorefname{~\ref{fig:rb_sr}} exhibits the prediction performance for different clusters. Our \name surpasses other baselines by a significant margin, particularly for less popular regions (marked by black circles), which is consistent with results in \figureautorefname{~\ref{fig:pred_error}}. This also verifies the robustness of \name to accurately predicts traffic flows of different types of spatial regions.

For temporal heterogeneity, according to urban traffic rhythms~\cite{wang2019understanding}, we partition a workday into four time periods and a holiday (weekend included) into two time periods, whose categories are given in \figureautorefname{~\ref{fig:rb_tc}}. \figureautorefname{~\ref{fig:rb_tr}} presents the evaluation performance. Our \name beats the baselines in terms of every category. Furthermore, \name shows a significant improvement in categories 3 and 5 that denote the nighttime of workdays and holidays. During these times, traffic flow data are typically sparse, making it difficult for baselines to produce accurate predictions. \name can handle this situation because we inject the temporal heterogeneity into the time-aware region embeddings.

% Table generated by Excel2LaTeX from sheet 'params'
% \begin{table}[t]\small
%     \centering
%     \setlength{\tabcolsep}{1mm}
%       \begin{tabular}{ccccc}
%       \toprule
%             & AGCRN & STSGCN & STFGNN & HeST \\
%       \midrule
%       NYCBike1 & 750k/675k & 208k/125k & 583k/524k & 196k/73k \\
%     NYCBike2 & 751k/676k & 213k/125k & 1.3m/1.2m & 365k/110k \\
%     NYCTaxi & 751k/676k & 213k/125k & 1.3m/1.2m & 365k/110k \\
%     BJTaxi & 759k/684k & 266k/125k & 33.7m/33.5m & 789k/535k \\
%     %   NYCBike1 & 750k(1:9.0) &       & 583k(1:8.9) & 196k(1:0.6) \\
%     %   NYCBike2 & 751k(1:9.0) &       & 1m(1:20.2) & 365k(1:0.4) \\
%     %   NYCTaxi & 751k(1:9.0) &       & 1m(1:20.2) & 365k(1:0.4) \\
%     %   BJTaxi & 759k(1:9.1) &       & 33m(1:289.1) & 789k(1:2.1) \\
%       \bottomrule
%       \end{tabular}%
%       %\vspace{-.2cm}
%       \caption{Comparison of parameter amount of different heterogeneity modeling methods. The amount of parameters introduced to model heterogeneity is listed after the total number of parameters and separated by a `/'.}%\vspace{-.3cm}
%     \label{tab:param}%
%   \end{table}%

% Next, we carry out an analysis on the parameter growth caused by heterogeneity modeling. As shown in \tablename~\ref{tab:param}, our \name introduce fewer parameters than other baselines. Among the baselines, STFGNN uses the most extra parameters, exceeding 95\% of the total parameters. In conclusion, our \name can model the spatial and temporal heterogeneity without significantly increasing the model parameter.

\subsection{Qualitative Study (RQ4)} 

\begin{figure}[t]
\centering
\includegraphics[width=0.95\columnwidth]{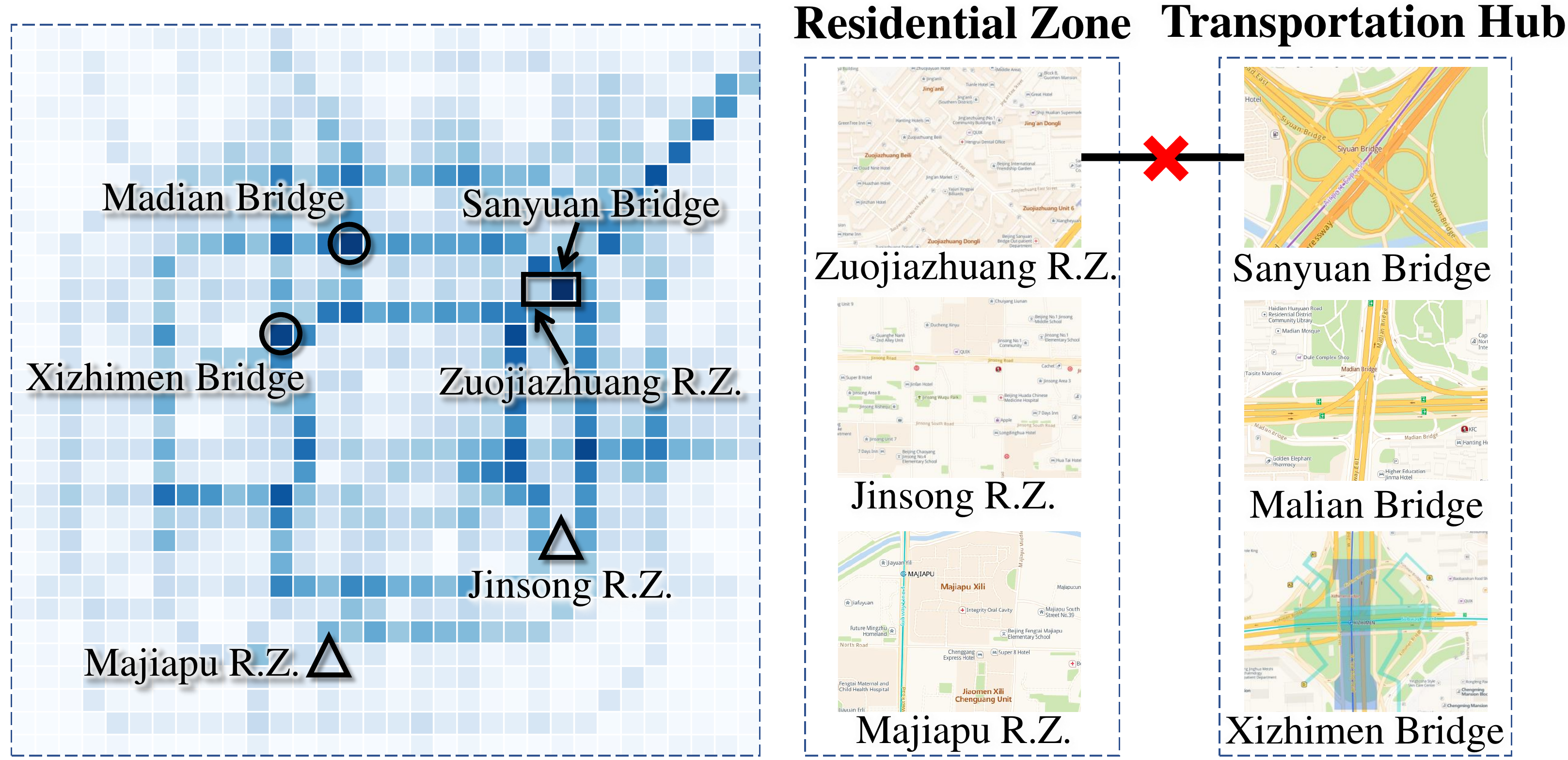}
% \vspace{-.2cm}
\caption{Case study on the adaptive graph augmentation.}
\vspace{-.3cm}
\label{fig:cs_adaaug}
\end{figure}

In \figureautorefname~\ref{fig:cs_adaaug}, we investigate the heterogeneity-guided graph topology-level augmentation on BJTaxi. Our augmentation method adaptively removes connections between adjacent regions with heterogeneous traffic patterns, \ie Zuojiazhuang Residential Zone and Sanyuan Bridge (a transportation hub). Meanwhile, it builds connections between distant regions with similar latent urban function, \eg Xizhimen Bridge and Sanyuan Bridge that are both transportation hubs. In this way, our \name can not only debias the region connections with low inter-correlated traffic patterns, but also capture the long-range region dependencies with the global urban context.  

\begin{figure}[t]
\centering
\includegraphics[width=0.80\columnwidth]{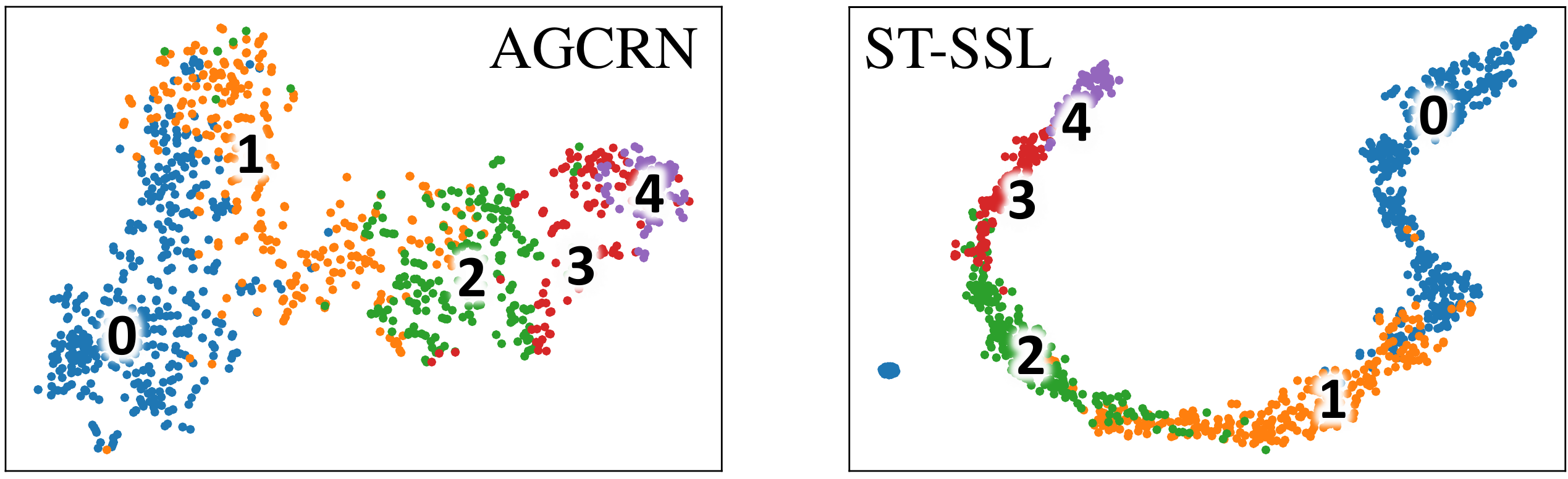}
% \vspace{-.2cm}
\caption{t-SNE visualization of embeddings on BJTaxi.}
\vspace{-.3cm}
\label{fig:cs}
\end{figure}

To further explore why the embeddings obtained by \name can deliver more accurate traffic prediction than AGCRN, we visualize them on BJTaxi by t-SNE~\cite{van2008visualizing}. We plot the learned embeddings of all regions with ground truth classes the same as \figureautorefname{~\ref{fig:rb_sc}}. As shown in \figureautorefname{~\ref{fig:cs}}, samples in the same class are more compact and those of different classes are significantly better separated for \name. This enables \name to be aware of spatial heterogeneity and transfer information between regions in the same class, which facilitates predictions.

\section{Related Work}\label{sec:related}

\subsubsection{Deep Learning for Traffic Prediction.} Many efforts have been devoted to developing traffic prediction techniques based on various neural networks. RNN~\cite{wang2019empowering, ji2020interpretable} and 1D CNN~\cite{wang2022traffic,wang2016traffic} are applied to capture the temporal dependencies in traffic series. CNN~\cite{zhang2017deep,yao2019revisiting}, GNN~\cite{zhang2020spatial,ji2022stden}, and attention mechanism~\cite{zheng2020gman} are introduced to incorporate the spatial information. However, most of them neglect the spatio-temporal heterogeneity problem. Recently, some works model the heterogeneity by using multiple models~\cite{yuan2018hetero} or multiple sets of parameters~\cite{bai2020adaptive,li2021spatial}, and some use meta learning to generate different weights based on static features of different regions~\cite{pan2019urban,ye2022meta}. However, these methods either introduce a number of parameters that may cause an overfitting problem or require external data that may be not available. To overcome these limitations, we incorporate self-supervised learning into traffic prediction to explore spatial and temporal heterogeneity.

\subsubsection{Self-Supervised Learning for Representation Learning.} Self-supervised learning aims to extract useful information from input data to improve the representation quality~\cite{hendrycks2019using}. 
The general paradigm is to augment the input data and then design pretext tasks as pseudo-labels for representation learning. 
It has achieved great success with text~\cite{kenton2019bert}, image~\cite{chen2020simple}, and graph data~\cite{wang2017community}.
% and audio data~\cite{oord2018representation}. 
Motivated by these works, we develop an adaptive data augmentation method for spatio-temporal graph data and introduce two pretext tasks to learn representations that are robust to spatio-temporal heterogeneity, which has not been well explored in existing traffic flow prediction methods.

\section{Conclusion and Future Work}

This work investigated the traffic prediction problem by proposing a novel spatio-temporal self-supervised learning (\name) framework. Specifically, we integrated temporal and spatial convolutions to encode spatial-temporal traffic patterns. Then, we devised $i$) a spatial self-supervised learning paradigm that consists of an adaptive graph augmentation and a clustering-based generative task, and $ii$) a temporal self-supervised learning paradigm that relies on a time-aware contrastive task, to supplement the main traffic flow prediction task with spatial and temporal heterogeneity-aware self-supervised signals. Comprehensive experiments on four traffic flow datasets demonstrated the robustness of \name. The future work lies in extending our spatial-temporal SSL framework to a model-agnostic paradigm.

% Comprehensive experiments on four datasets demonstrate that the proposed \name consistently outperforms 8 baselines over different types of regions and time periods.
% This work investigated the spatial and temporal heterogeneity in traffic flow prediction by proposing a novel Heterogeneity-aware Spatio-Temporal (\name) framework based on a self-supervised learning paradigm. Specifically, we integrated temporal and spatial convolutions to encode spatial-temporal traffic patterns. Then, the adaptive heterogeneity-guided data augmentation was performed over the traffic flow graph data at both traffic data-level and graph structure-level. On top of the augmented traffic graph, we constructed a clustering-based generative task and a time-aware contrastive task to supplement the main traffic flow prediction task with spatial and temporal heterogeneity-aware self-supervised signals. The future work lies in extending our framework to a model-agnostic paradigm.

% In \name, we designed two self-supervised tasks to learn the spatial and temporal heterogeneity respectively, and developed an adaptive augmentation method to prepare data for the tasks. Comprehensive experiments on four datasets demonstrate that the proposed \name consistently outperforms 8 baselines over different types of regions and time periods. The future work lies in extending our framework to a model-agnostic paradigm.

\clearpage

\section*{Acknowledgments}

This work was supported by the National Key R\&D Program of China (2019YFB2101804). Prof. Wang’s work was supported by the National Natural Science Foundation of China (No. 72222022, 82161148011, 72171013), the Fundamental Research Funds for the Central Universities (YWF-22-L-838) and the DiDi Gaia Collaborative Research Funds. Dr. Zhang’s work was supported by the National Natural Science Foundation of China (No. 62172034) and the Beijing Nova Program (Z201100006820053).

\bibliography{base}

\begin{thebibliography}{28}
\providecommand{\natexlab}[1]{#1}

\bibitem[{Bai et~al.(2020)Bai, Yao, Li, Wang, and Wang}]{bai2020adaptive}
Bai, L.; Yao, L.; Li, C.; Wang, X.; and Wang, C. 2020.
\newblock Adaptive graph convolutional recurrent network for traffic forecasting.
\newblock \emph{NeurIPS}, 33: 17804--17815.

\bibitem[{Castro-Neto et~al.(2009)Castro-Neto, Jeong, Jeong, and Han}]{castro2009online}
Castro-Neto, M.; Jeong, Y.-S.; Jeong, M.-K.; and Han, L.~D. 2009.
\newblock Online-SVR for short-term traffic flow prediction under typical and atypical traffic conditions.
\newblock \emph{Expert Systems with Applications}, 36(3): 6164--6173.

\bibitem[{Chen et~al.(2020)Chen, Kornblith, Norouzi, and Hinton}]{chen2020simple}
Chen, T.; Kornblith, S.; Norouzi, M.; and Hinton, G. 2020.
\newblock A simple framework for contrastive learning of visual representations.
\newblock In \emph{ICML}, 1597--1607.

\bibitem[{Hendrycks et~al.(2019)Hendrycks, Mazeika, Kadavath, and Song}]{hendrycks2019using}
Hendrycks, D.; Mazeika, M.; Kadavath, S.; and Song, D. 2019.
\newblock Using self-supervised learning can improve model robustness and uncertainty.
\newblock \emph{NeurIPS}, 32.

\bibitem[{Ji et~al.(2022)Ji, Wang, Jiang, Jiang, and Zhang}]{ji2022stden}
Ji, J.; Wang, J.; Jiang, Z.; Jiang, J.; and Zhang, H. 2022.
\newblock {STDEN}: Towards physics-guided neural networks for traffic flow prediction.
\newblock In \emph{AAAI}, volume~36, 4048--4056.

\bibitem[{Ji et~al.(2020)Ji, Wang, Jiang, Ma, and Zhang}]{ji2020interpretable}
Ji, J.; Wang, J.; Jiang, Z.; Ma, J.; and Zhang, H. 2020.
\newblock Interpretable spatiotemporal deep learning model for traffic flow prediction based on potential energy fields.
\newblock In \emph{ICDM}, 1076--1081.

\bibitem[{Kenton and Toutanova(2019)}]{kenton2019bert}
Kenton, J. D. M.-W.~C.; and Toutanova, L.~K. 2019.
\newblock BERT: Pre-training of deep bidirectional transformers for language understanding.
\newblock In \emph{Proceedings of the 2019 Conference of the North American Chapter of the Association for Computational Linguistics}, 4171--4186.

\bibitem[{Kumar and Vanajakshi(2015)}]{kumar2015short}
Kumar, S.~V.; and Vanajakshi, L. 2015.
\newblock Short-term traffic flow prediction using seasonal ARIMA model with limited input data.
\newblock \emph{European Transport Research Review}, 7(3): 1--9.

\bibitem[{Li and Zhu(2021)}]{li2021spatial}
Li, M.; and Zhu, Z. 2021.
\newblock Spatial-temporal fusion graph neural networks for traffic flow forecasting.
\newblock In \emph{AAAI}, volume~35, 4189--4196.

\bibitem[{Pan et~al.(2019{\natexlab{a}})Pan, Liang, Wang, Yu, Zheng, and Zhang}]{pan2019urban}
Pan, Z.; Liang, Y.; Wang, W.; Yu, Y.; Zheng, Y.; and Zhang, J. 2019{\natexlab{a}}.
\newblock Urban traffic prediction from spatio-temporal data using deep meta learning.
\newblock In \emph{ACM SIGKDD}, 1720--1730.

\bibitem[{Pan et~al.(2019{\natexlab{b}})Pan, Wang, Wang, Yu, Zhang, and Zheng}]{pan2019matrix}
Pan, Z.; Wang, Z.; Wang, W.; Yu, Y.; Zhang, J.; and Zheng, Y. 2019{\natexlab{b}}.
\newblock Matrix factorization for spatio-temporal neural networks with applications to urban flow prediction.
\newblock In \emph{CIKM}, 2683--2691.

\bibitem[{Song et~al.(2020)Song, Lin, Guo, and Wan}]{song2020spatial}
Song, C.; Lin, Y.; Guo, S.; and Wan, H. 2020.
\newblock Spatial-temporal synchronous graph convolutional networks: A new framework for spatial-temporal network data forecasting.
\newblock In \emph{AAAI}, volume~34, 914--921.

\bibitem[{Van~der Maaten and Hinton(2008)}]{van2008visualizing}
Van~der Maaten, L.; and Hinton, G. 2008.
\newblock Visualizing data using t-SNE.
\newblock \emph{Journal of machine learning research}, 9(11).

\bibitem[{Wang et~al.(2016)Wang, Gu, Wu, Liu, and Xiong}]{wang2016traffic}
Wang, J.; Gu, Q.; Wu, J.; Liu, G.; and Xiong, Z. 2016.
\newblock Traffic speed prediction and congestion source exploration: A deep learning method.
\newblock In \emph{ICDM}, 499--508.

\bibitem[{Wang et~al.(2022)Wang, Ji, Jiang, and Sun}]{wang2022traffic}
Wang, J.; Ji, J.; Jiang, Z.; and Sun, L. 2022.
\newblock Traffic Flow Prediction Based on Spatiotemporal Potential Energy Fields.
\newblock \emph{IEEE Transactions on Knowledge and Data Engineering}, 1--14.

\bibitem[{Wang et~al.(2021)Wang, Jiang, Jiang, Li, and Zhao}]{wang2021libcity}
Wang, J.; Jiang, J.; Jiang, W.; Li, C.; and Zhao, W.~X. 2021.
\newblock {LibCity}: An open library for traffic prediction.
\newblock In \emph{Proceedings of the 29th International Conference on Advances in Geographic Information Systems}, 145--148.

\bibitem[{Wang et~al.(2019{\natexlab{a}})Wang, Wu, Wang, Gao, and Xiong}]{wang2019understanding}
Wang, J.; Wu, J.; Wang, Z.; Gao, F.; and Xiong, Z. 2019{\natexlab{a}}.
\newblock Understanding urban dynamics via context-aware tensor factorization with neighboring regularization.
\newblock \emph{IEEE Transactions on Knowledge Data Engineering}, 32(11): 2269--2283.

\bibitem[{Wang et~al.(2019{\natexlab{b}})Wang, Wu, Zhao, Peng, and Lin}]{wang2019empowering}
Wang, J.; Wu, N.; Zhao, W.~X.; Peng, F.; and Lin, X. 2019{\natexlab{b}}.
\newblock Empowering {A*} search algorithms with neural networks for personalized route recommendation.
\newblock In \emph{ACM SIGKDD}, 539--547.

\bibitem[{Wang et~al.(2017)Wang, Cui, Wang, Pei, Zhu, and Yang}]{wang2017community}
Wang, X.; Cui, P.; Wang, J.; Pei, J.; Zhu, W.; and Yang, S. 2017.
\newblock Community preserving network embedding.
\newblock In \emph{Proceedings of the AAAI conference on artificial intelligence}, volume~31.

\bibitem[{Yao et~al.(2019)Yao, Tang, Wei, Zheng, and Li}]{yao2019revisiting}
Yao, H.; Tang, X.; Wei, H.; Zheng, G.; and Li, Z. 2019.
\newblock Revisiting spatial-temporal similarity: A deep learning framework for traffic prediction.
\newblock In \emph{AAAI}, volume~33, 5668--5675.

\bibitem[{Ye et~al.(2022)Ye, Fang, Sun, Zhang, and Xiang}]{ye2022meta}
Ye, X.; Fang, S.; Sun, F.; Zhang, C.; and Xiang, S. 2022.
\newblock Meta graph transformer: A novel framework for spatial-temporal traffic prediction.
\newblock \emph{Neurocomputing}, 491: 544--563.

\bibitem[{Yu, Yin, and Zhu(2018)}]{yu2018spatio}
Yu, B.; Yin, H.; and Zhu, Z. 2018.
\newblock Spatio-temporal graph convolutional networks: a deep learning framework for traffic forecasting.
\newblock In \emph{IJCAI}, 3634--3640.

\bibitem[{Yuan, Zhou, and Yang(2018)}]{yuan2018hetero}
Yuan, Z.; Zhou, X.; and Yang, T. 2018.
\newblock {Hetero-ConvLSTM}: A deep learning approach to traffic accident prediction on heterogeneous spatio-temporal data.
\newblock In \emph{ACM SIGKDD}, 984--992.

\bibitem[{Zhang, Zheng, and Qi(2017)}]{zhang2017deep}
Zhang, J.; Zheng, Y.; and Qi, D. 2017.
\newblock Deep spatio-temporal residual networks for citywide crowd flows prediction.
\newblock In \emph{AAAI}, volume~31, 1655--1661.

\bibitem[{Zhang et~al.(2020)Zhang, Huang, Xu, and Xia}]{zhang2020spatial}
Zhang, X.; Huang, C.; Xu, Y.; and Xia, L. 2020.
\newblock Spatial-temporal convolutional graph attention networks for citywide traffic flow forecasting.
\newblock In \emph{Proceedings of the 29th ACM International Conference on Information and Knowledge Management}, 1853--1862.

\bibitem[{Zhang et~al.(2021)Zhang, Huang, Xu, Xia, Dai, Bo, Zhang, and Zheng}]{zhang2021traffic}
Zhang, X.; Huang, C.; Xu, Y.; Xia, L.; Dai, P.; Bo, L.; Zhang, J.; and Zheng, Y. 2021.
\newblock Traffic flow forecasting with spatial-temporal graph diffusion network.
\newblock In \emph{AAAI}, volume~35, 15008--15015.

\bibitem[{Zheng et~al.(2020)Zheng, Fan, Wang, and Qi}]{zheng2020gman}
Zheng, C.; Fan, X.; Wang, C.; and Qi, J. 2020.
\newblock {GMAN}: A graph multi-attention network for traffic prediction.
\newblock In \emph{AAAI}, volume~34, 1234--1241.

\bibitem[{Zhu et~al.(2021)Zhu, Xu, Yu, Liu, Wu, and Wang}]{zhu2021graph}
Zhu, Y.; Xu, Y.; Yu, F.; Liu, Q.; Wu, S.; and Wang, L. 2021.
\newblock Graph contrastive learning with adaptive augmentation.
\newblock In \emph{Proceedings of the Web Conference 2021}, 2069--2080.

\end{thebibliography}

\end{document}